\documentclass[letterpaper, 10 pt, conference]{ieee/ieeeconf}  

\IEEEoverridecommandlockouts                              
\usepackage{graphics} 
\usepackage{epsfig} 
\usepackage{mathptmx} 
\usepackage{times} 
\usepackage{amsmath} 
\usepackage{amssymb}  
\usepackage{xcolor}
\usepackage{color}
\usepackage[linesnumbered,ruled,vlined]{algorithm2e}
\usepackage{algorithmic}
\usepackage{subfigure}
\usepackage{multirow}
\usepackage{float}
\usepackage{algorithm2e}
\usepackage{booktabs}
\usepackage[utf8x]{inputenc}
\usepackage{wrapfig}
\usepackage{threeparttable}
\usepackage{boldline}

\newcommand{\YOON}[1] {
	\textcolor{blue}{\bfseries{YOON: {#1}}}
}
\newcommand{\MC}[1] {
	\textcolor{red}{\bfseries{MC: {#1}}}
}
\newcommand{\DHK}[1] {
	\textcolor{purple}{\bfseries{DH: {#1}}}
}
\newcommand{\HC}[1] {
	\textcolor{cyan}{\bfseries{HC: {#1}}}
}

\DeclareMathAlphabet{\mathcal}{OMS}{cmsy}{m}{n}
\renewcommand{\baselinestretch}{1}

\SetKwComment{Comment}{$\triangleright$\ }{}

\title{\LARGE \bf
	TORM: Fast and Accurate Trajectory Optimization of \\
	Redundant Manipulator given an End-Effector Path
}

\author{Mincheul Kang$^{1}$, Heechan Shin$^{1}$, Donghyuk Kim$^{1}$ and Sung-Eui Yoon$^{2}$
\thanks{{$^1$}Mincheul Kang ({\tt\small mincheul.kang@kaist.ac.kr}),
{$^1$}Heechan Shin ({\tt\small shin\_heechan@kaist.ac.kr}) and
{$^1$}Donghyuk Kim ({\tt\small donghyuk.kim@kaist.ac.kr}) are with the School of Computing, and
{$^2$}Sung-Eui Yoon (Corresponding author, {\tt\small sungeui@kaist.edu}) is with the Faculty of School of Computing, KAIST at Daejeon, Korea 34141}
}

\newcommand{\Skip}[1]{}
\renewcommand{\paragraph}[1]{{\bf {#1}}}  

\def\HiLi{\leavevmode\rlap{\hbox to 
		\hsize{\color{yellow!50}\leaders\hrule height .8\baselineskip depth .5ex\hfill}}}

\begin{document}
	\maketitle
	\thispagestyle{empty}
	\pagestyle{empty}
	
	\newtheorem{thm}{Theorem}
	\newtheorem{lem}[thm]{Lemma}
	\newtheorem{col}[thm]{Corollary}
	
	\begin{abstract}
A redundant manipulator has multiple inverse kinematics solutions per end-effector pose. 
Accordingly, there can be many trajectories for joints that follow a given end-effector path in the Cartesian space.
In this paper, we present a trajectory optimization of a redundant manipulator (TORM) to synthesize a trajectory that follows a given end-effector path accurately, while achieving smoothness and collision-free manipulation.
Our method holistically incorporates three desired properties
into the trajectory optimization process by integrating the
Jacobian-based inverse kinematics solving method and an optimization-based motion planning approach.
Specifically, we optimize a trajectory using two-stage gradient descent to
	reduce potential competition between different properties during the
	update.
To avoid falling into local minima, we iteratively explore different candidate trajectories with our local update.
We compare our method with state-of-the-art methods in test scenes including external obstacles and two non-obstacle problems.  Our method robustly minimizes the pose error in a progressive manner while
	satisfying various desirable properties. 
\end{abstract}

	\section{INTRODUCTION}
\label{sec:1}

Remote control of various robots has been 
one of the main challenges in the robotics, while it is commonly used for cases 
where it is difficult or dangerous for a human 
to perform 
tasks~\cite{katyal2014approaches, long2019optimization}.
In this remote control scenario, 
a robot has to follow the task command accurately while considering its surrounding environment and constraints of the robot itself. 

In the case of a redundant manipulator that this paper focuses on, a sequence of finely-specified joint configurations is required to follow the end-effector path in a Cartesian space accurately.
Traditionally, inverse kinematics (IK) has been used to determine joint configurations given an end-effector pose. 
The traditional IK, however, cannot consider the continuity of configurations, collision avoidance, and kinematic singularities that arise when considering to follow the end-effector path.

Path-wise IK approaches~\cite{rakita2018relaxedik, rakita2019stampede} solve this problem using non-linear optimization by considering aforementioned constraints. 
These approaches avoid self-collisions using a neural network, but they do not deal with collisions for external obstacles.
On the other hand, prior methods~\cite{holladay2016distance, holladay2019minimizing} based on motion planning approach consider external obstacles and use IK solutions to synthesize a trajectory that is following the desired path in Cartesian space.
By simply using IK solutions, however, these methods have time or structural difficulties in getting a highly-accurate solution (Sec.~\ref{sec:2_motion_planning}).

\begin{figure}[t]
	\vspace{0.3cm}
	\centering
	\includegraphics[width=\columnwidth]{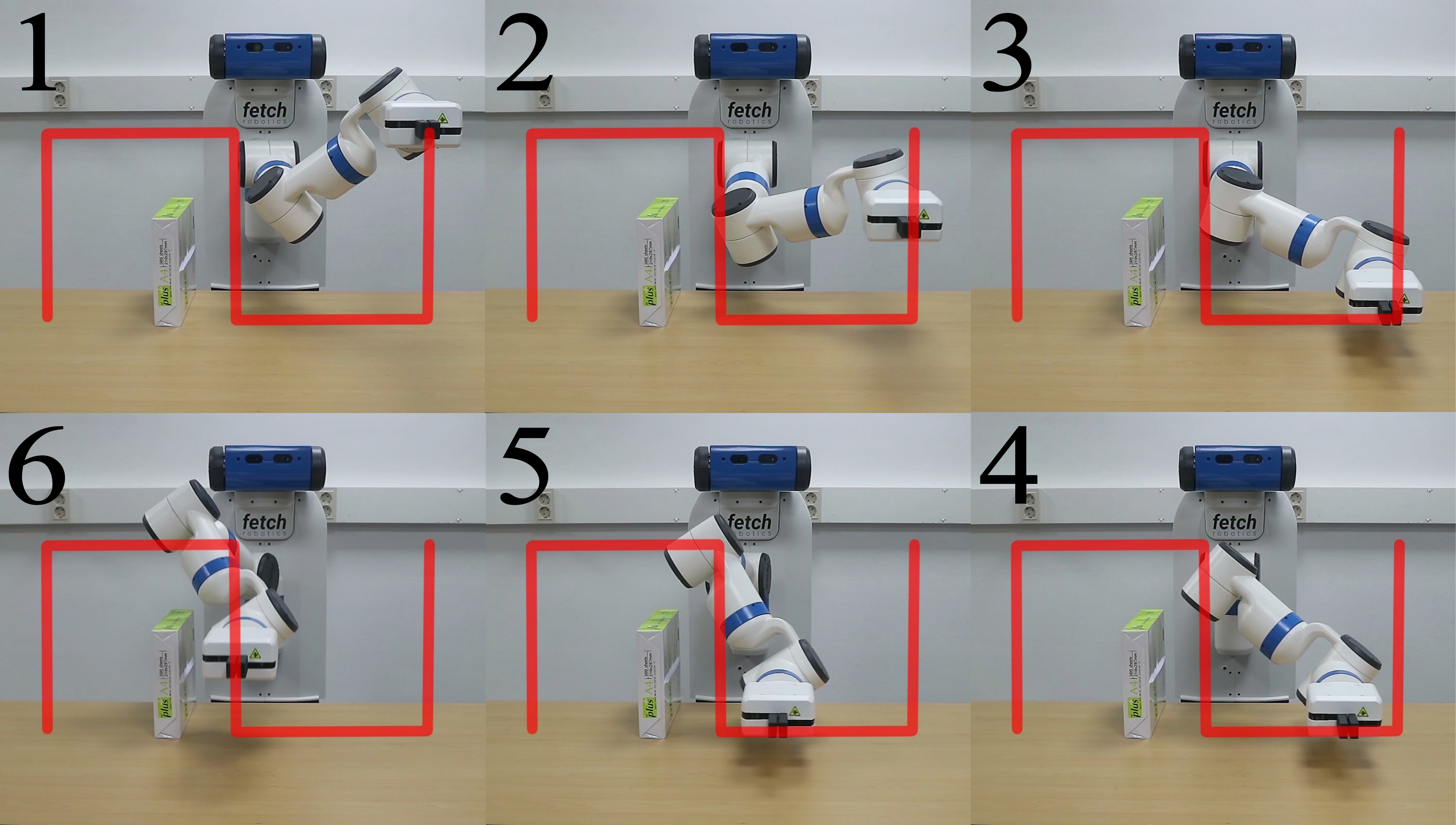}
	\caption{
	    These figures show a sequence of maneuvering the Fetch manipulator to follow the specified end-effector path (red lines).
	    Our method generates the trajectory that accurately follows the given end-effector path, while 
	    avoiding obstacles such as the pack of A4 paper and the table.
    }
	\label{fig:main}
	\vspace{-0.33cm}
\end{figure}

\noindent
\textbf{Main Contributions.}
In this work, we present a trajectory optimization of a redundant manipulator (TORM) 
for synthesizing a trajectory that is accurately following a given path as well as smooth and 
collision-free against the robot itself and external obstacles (Fig.~\ref{fig:main}).
Our method incorporates these properties into the optimization process by
holistically 
integrating the Jacobian-based IK solving method and an optimization-based
motion planning approach
(Sec.~\ref{sec:3_objective_function}).
For the effective optimization, we consider different properties of our
objectives through a two-stage update manner, which alternates
making a constraint-satisfying trajectory and following a given end-effector
path accurately (Sec.~\ref{sec:3_optimization_process}).  
To avoid local minima within our optimization process, we perform iterative
exploration by considering other alternative trajectories and progressively
identifying a better trajectory
(Sec.~\ref{sec:3_iterative_local_search}).

To compare our method with the state-of-the-art methods,
RelaxedIK~\cite{rakita2018relaxedik}, Stampede~\cite{rakita2019stampede} and
the work of Holladay et al.~\cite{holladay2019minimizing}, we test two scenes with external obstacles and two non-obstacle problems
(Sec.~\ref{sec:4_exp_setting}).
Overall, we observe that our method achieves more accurate solutions given
the equal planning time over the tested prior methods.
Also, our method robustly minimizes the pose error reasonably fast with the
anytime property, which quickly finds an initial trajectory and refines the
trajectory~\cite{karaman2011anytime} (Sec.~\ref{sec:4_evaluation}).  This
result is mainly resulted by the holistic optimization process.
The synthesized results of tested problems and real robot verifications can be
seen in the attached video.

	\section{RELATED WORK}
\label{sec:2}

In this section, we discuss prior studies in the fields of inverse kinematics 
and motion planning for following 
the desired end-effector path.

\subsection{Inverse Kinematics}
\label{sec:2_inverse_kinematics}

Inverse kinematics (IK) has been studied widely to find a joint configuration 
from an end-effector pose \cite{buss2004introduction}.  
In the case of a redundant manipulator, where our
target robots belong to, there can be multiple joint configurations from a
single end-effector pose.  For this problem, several techniques for quickly
finding solutions have been
proposed~\cite{diankov2010automated, sinha2019searchik}.
In particular, task-priority
IK algorithms~\cite{chiacchio1991closed, chiaverini1997singularity,
kanoun2011kinematic}
prioritize solutions based on an objective function for each purpose, e.g.,
kinematic singularity or constraint task.

Most methods that synthesize a trajectory geometrically constrained for an
end-effector pose use the Jacobian matrix 
for finding a feasible solution~\cite{stilman2007task, berenson2009manipulation, vahrenkamp2009humanoid, kunz2012manipulation}.
Unfortunately, the Jacobian matrix is the first derivative of the vector-valued
function with multiple variables and thus 
it can cause false-negative failures
by getting stuck in local minima or
convergence failures due to the limits of the joint angle~\cite{beeson2015trac}.

Trac-IK~\cite{beeson2015trac} points out the problem and improves success rates by using randomly selected joint configurations
and sequential quadratic
programming. Nonetheless, its solutions do not guarantee continuity of a sequence of joint configurations to make a feasible trajectory~\cite{rakita2018relaxedik}.
In summary, the traditional IK approaches have different strengths and weaknesses for synthesizing a feasible trajectory.

To get alleviated solutions, 
many studies have present optimization techniques with objective functions for synthesizing a feasible trajectory with matching end-effector poses.  
Luo and Hauser~\cite{luo2012interactive} use a geometric and temporal optimization to generate a dynamically-feasible trajectory from a sketch input. 
Recently, Rakita et al.~\cite{rakita2018relaxedik} proposed a real-time approach using a weighted-sum non-linear optimization, called RelaxedIK, to solve the IK
problem for a sequence of motion inputs.  
Since the collision checking has a relatively large computational overhead, RelaxedIK uses a neural network for fast self-collision avoidance.
 
Based on RelaxedIK, two studies~\cite{rakita2019stampede, praveena2019user} are proposed for synthesizing a
highly-accurate trajectory on off-line. 
One of them is Stampede~\cite{rakita2019stampede}, which finds an optimal trajectory using a dynamic programming algorithm in a discrete-space-graph that is built by samples of IK solutions.  
The other work~\cite{praveena2019user} generates multiple candidate trajectories from multiple starting configurations and then selects the best trajectory with a user guide by allowing a deviation if there are risks of self-collisions or kinematic singularities~\cite{praveena2019user}. 

The aforementioned methods, called path-wise IK methods, optimize joint
configurations at finely divided end-effector poses.  
These optimization methods synthesize an accurate and feasible trajectory
satisfied with several constraints, i.e., continuity of configurations,
collision avoidance, and kinematic singularities.  Inspired by this strategy,
we propose a trajectory optimization of a redundant manipulator (TORM) to get a
collision-free and highly-accurate solution.  Unlike prior path-wise IK works,
however, our method considers self-collision as well as external obstacles,
thanks to 
a tight integration of an efficient
collision avoidance method using a signed distance field.


\Skip{
	Our method also seeks to find a highly-accurate solution for the
	redundant manipulators while avoiding self-collisions.
	On top of that, we additionally consider collision avoidance against external
	obstacles\YOON{it does not come across as novelty}.
	To synthesize an accurate and feasible trajectory without any collisions against a robot itself as well as obstacles, we propose a trajectory optimization of a redundant manipulator (TORM) 
	that extends the
	CHOMP~\cite{zucker2013chomp}, one of the most prominent
	optimization-based motion planning approaches,
	 for our problem. 
}

\subsection{Motion Planning for Following an End-effector Path} 
\label{sec:2_motion_planning}

Motion planning involves a collision-free trajectory from a start configuration to a goal configuration.
Based on the framework of motion planning, Holladay et al.~\cite{holladay2016distance, holladay2019minimizing} present two methods to follow the desired end-effector path in Cartesian space.
One~\cite{holladay2016distance} uses an optimization-based method, specifically
Trajopt~\cite{schulman2013finding}, by applying the discrete Fr\'{e}chet
distance that approximately measures the similarity of two curves.
Although the optimization-based motion planning approaches quickly find a
collision-free trajectory using efficient collision avoidance methods, these
approaches can be stuck in local minima due to several constraints (e.g., joint
limits and collisions).
To assist its optimizer, this approach separately plans a trajectory by
splitting the end-effector path as a set of waypoints and then sampling an IK solution at each pose.

Its subsequent work~\cite{holladay2019minimizing} points out the limitation of
the prior work that is sampling only one IK solution for each pose.
Considering this property, it presents a sampling-based approach that
iteratively updates a layered graph by sampling new waypoints and IK solutions.
However, this method needs a lot of planning time to get a highly-accurate
solution, even with
lazy collision checking to reduce the
computational overhead.

Even though these methods find a collision-free trajectory by utilizing a
motion planning approach and random IK solutions, it is hard to get a
highly-accurate solution due to time or structural constraints.  To overcome
these difficulties, our method incorporates the Jacobian-based IK solving
method into our optimization process, instead of using only IK solutions.
The aforementioned path-wise IK approaches also use the objective function
to minimize the end-effector pose error, but do not combine it with the
objective function to avoid external obstacles.
On the other hand, our approach holistically integrates these different
objectives and constraints within an optimization framework, inspired by an
optimization-based motion planning approach, CHOMP~\cite{zucker2013chomp}, and
effectively computes refined trajectories based on our two-stage gradient
descent. 

	\section{Trajectory optimization}
\label{sec:3}

The problem we aims to solve is finding a trajectory, $\xi$, of a redundant manipulator for a given end-effector path, $\mathcal{X}$. 
The trajectory $\xi$ is a sequence of joint configurations, and the end-effector path, $\mathcal{X} \subset \mathbb{R}^{6}$, is defined in the 6-dimensional Cartesian space.

\begin{figure}[t]
	\vspace{0.25cm}
	\centering
	\includegraphics[width=2.5in]{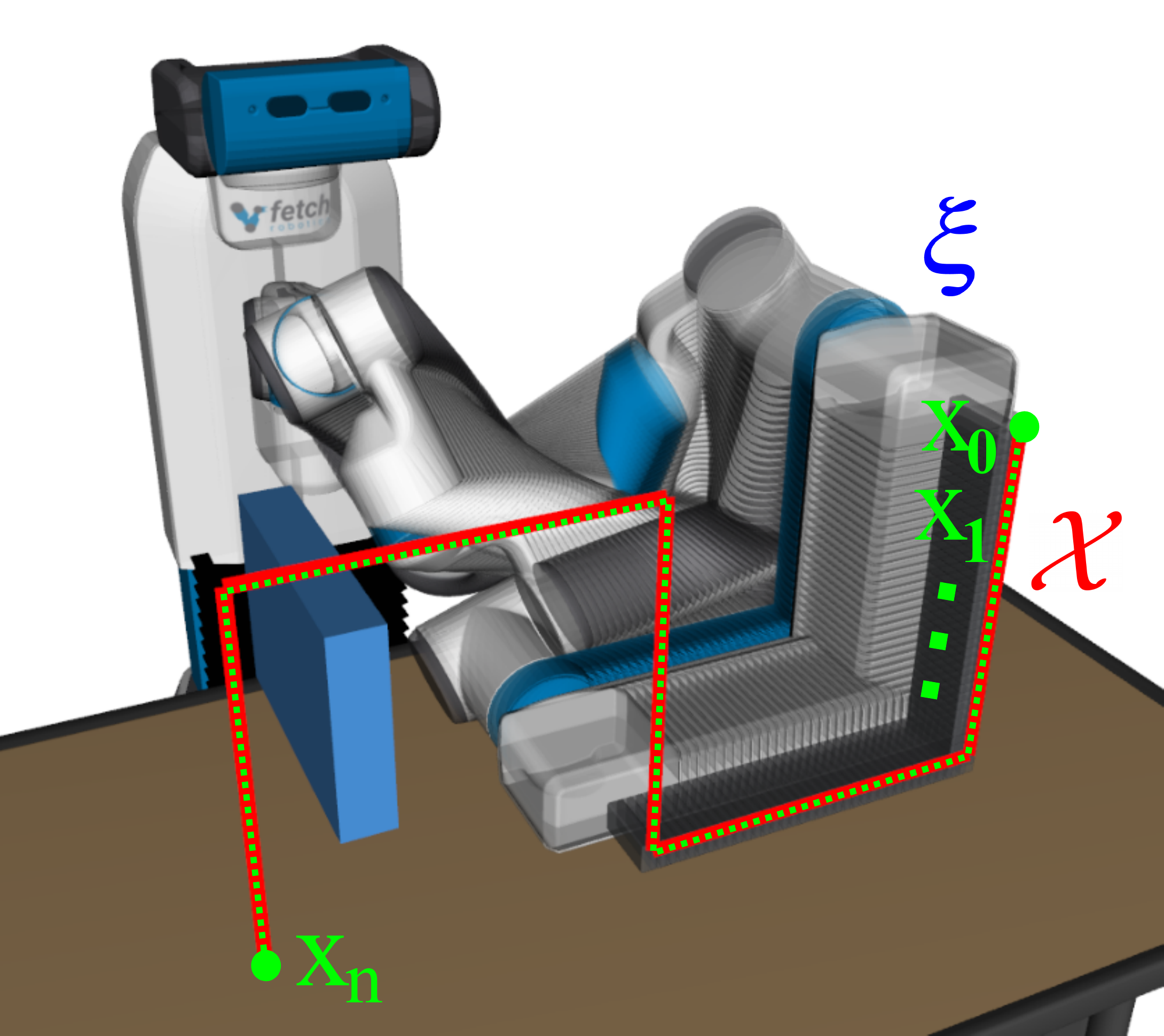}
	\caption{
		This figure shows our problem that is synthesizing a feasible and
		accurate trajectory $\xi$ for a given end-effector path $\mathcal{X}$. The red line is $\mathcal{X}$, which is approximated by end-effector poses $\widetilde{\mathcal{X}} = (\mathrm{x}_0, \mathrm{x}_1, ..., \mathrm{x}_{n})$ (green dots).
		The trajectory $\xi$ is computed at end-effector poses $\widetilde{\mathcal{X}}$.
		The part of the synthesized trajectory shows that the
		end-effector follows the red line, and the sequence of joint
		configurations is smooth and collision-free while avoiding
		obstacles such as the blue box and the table.
	}
	\label{fig:problem}
	\vspace{-0.13cm}
\end{figure}

We solve the problem by applying the waypoint parameterization~\cite{flash1988communication} of the path that finely splits the given end-effector path, $\mathcal{X} \approx \widetilde{\mathcal{X}} = (\mathrm{x}_0, \mathrm{x}_1, ..., \mathrm{x}_{n})$; $\mathrm{x}$ is an end-effector pose.
We then compute the joint configurations for end-effector poses $\widetilde{\mathcal{X}}$.
As a result, the trajectory is approximated as: 
$\xi \approx (q_0^T,
q_1^T, ..., q_{n}^T)^T \subset \mathbb{R}^{(n+1) \times d}$,
where $d$ is the degree of freedom (DoF) of a manipulator; $q_0$ is a start configuration and $q_n$ is a goal configuration for the computed trajectory of joint configurations.
Main notations are summarized in Table~\ref{tab:notations}.

Our target robot is a redundant manipulator that has multiple joint
configurations given an end-effector pose; if $d > 6$, it has infinitely many valid solutions.
Among many candidates, we quickly optimize a set of joint configurations as a trajectory to follow the given end-effector path accurately, while avoiding collisions for external obstacles and the robot itself (Fig.~\ref{fig:problem}).

\Skip{
Among many candidates, we find appropriate joint configurations to generate a trajectory that is smooth and accurate as well as collision-free for the given
end-effector path (Fig.~\ref{fig:problem}). 
}

To achieve our goal, we present a trajectory optimization of a redundant
manipulator (TORM), inspired by an optimization-based motion planning approach,
specifically,  CHOMP~\cite{zucker2013chomp}, for finding a collision-free
trajectory from the start to the goal configurations.  Since CHOMP not only
quickly optimizes a trajectory using gradient descent, but also efficiently
avoids collisions by incorporating the objective function, we choose CHOMP as a
base trajectory optimization method of our approach.

\Skip{In the following section, we explain our optimization process including our objective function and update rule. We also introduce the quantum annealing to set a good initial trajectory and to avoid local minima.
}

Overall, we first define our own objective function, which has a new path
following objective (Sec.~\ref{sec:3_objective_function}). We repeatedly
generate different trajectories (Sec.~\ref{sec:3_iterative_local_search}),
which are refined by minimizing our objective function, based on our two-stage
gradient descent
(Sec.~\ref{sec:3_optimization_process}). 

\subsection{Objective Function}
\label{sec:3_objective_function}

We solve our problem by modeling an objective function holistically integrating
three different
properties that are 1) matching the end-effector pose, 2) avoiding collisions,
and 3) achieving smoothness:
\begin{equation}
\mathcal{U}(\xi) = \mathcal{F}_{pose}(\xi) + \lambda_1 \mathcal{F}_{obs}(\xi) + \lambda_2 \mathcal{F}_{smooth}(\xi),
\label{eq:objective_function}
\end{equation}
where $\lambda$ is a regularization constant.
$\mathcal{F}_{pose}$ is introduced to minimize the pose error between the target and current end-effector poses, and is defined using the squared Euclidean distance:
\begin{equation}
\mathcal{F}_{pose}(\xi) = \frac{1}{2}\sum_{i=0}^{n} \left\| \mathrm{x}_i - FK(q_i) \right\|^2,
\label{eq:objective_pose}
\end{equation}
where 
$FK(q_i)$ computes the end-effector pose at the joint configurations $q_i$
using forward kinematics (FK).

$\mathcal{F}_{obs}$ quantitatively measures proximity to external obstacles
using a signed distance field that can be calculated from the
geometry of workspace obstacles.
The robot body is simplified into spheres, which serve as conservative
bounding volumes for efficient computation. 
Overall, $\mathcal{F}_{obs}$ can be calculated highly fast and is formulated as:
\begin{equation}
\begin{aligned}
\mathcal{F}_{obs}(\xi) = & \sum_{i=0}^{n-1} \sum_{p}^{\mathcal{P}} \frac{1}{2} \left( c( x(q_{i+1}, p) ) + c( x(q_i, p) ) \right) \cdot \\
& \left\| x(q_{i+1}, p) - x(q_i, p) \right\| , 
\end{aligned}
\label{eq:objective_obs}
\end{equation}
where $x(q_i, p)$ is partial forward kinematics, i.e., a position of a body point $p \in \mathcal{P}$ in the workspace at a configuration $q_i$ and $c(\cdot)$ is an obstacle cost computed from the signed distance field.
At a high level, it approximately measures the sum of penetration depths
between the robot body and the obstacles.

$\mathcal{F}_{smooth}$ measures dynamical quantities, i.e., the squared sum of derivatives to encourage the smoothness between joint configurations:
\begin{equation}
\mathcal{F}_{smooth}(\xi) = \frac{1}{2}\sum_{i=0}^{n-1} \left\|\frac{q_{i+1}-q_{i}}{\Delta t}\right\|^2 .
\end{equation}


\setlength{\textfloatsep}{18pt}
\begin{table}[t]
	\vspace{0.4cm}
	\renewcommand \arraystretch{1.3}
	\begin{center}
		\caption{Notation table}
		\label{tab:notations}
		\begin{tabular}{|c|l|} 
			\hline
			\textbf{~Notation~} & \textbf{Description} \\
			\hline
			\multirow{2}{*}{$\widetilde{\mathcal{X}}$} & Target end-effector poses that are finely divided from \\
			~ & the given end-effector path $\mathcal{X} \subset \mathbb{R}^{6}$ \\
			\hline
			$\xi$ & Set of joint configurations corresponding to $\widetilde{\mathcal{X}}$ \\ 
			\hline
			\multirow{2}{*}{$q_i$} & Joint configuration on the trajectory $\xi \subset \mathbb{R}^{d}$ \\
			~ & at $i$-th end-effector pose $\mathrm{x}_i$ \\
			\hline
			$J$ & Jacobian matrix, i.e., $\frac{dx}{dq} \in \mathbb{R}^{6 \times d}$ \\
			\hline
			\multirow{2}{*}{$\mathcal{P}$} & Set of body points in the workspace approximating \\ ~ & the geometric shape of the manipulator \\
			\hline
			\multirow{2}{*}{$x(q, p)$} & Partial forward kinematics from the manipulator base to \\ ~ & a body point $p \in \mathcal{P}$ at a configuration $q$ \\ 
			\hline
		\end{tabular}
	\end{center}
	\vspace{-0.6cm}
\end{table}

\subsection{Two-Stage Gradient Descent Update}
\label{sec:3_optimization_process}

We iteratively update the trajectory to minimize the cost of the objective function consisting of three different terms.
To minimize the cost, our update rule is based on the gradient descent technique adopted in many optimization methods.
The functional gradient of the obstacle term, $\nabla \mathcal{F}_{obs}$, 
pushes the robot out of obstacles, and 
the functional gradient of the smooth term, $\nabla \mathcal{F}_{smooth}$, keeps the continuity between joint configurations. 

Both $\nabla \mathcal{F}_{obs}$ and $\nabla \mathcal{F}_{smooth}$ are responsible for synthesizing a feasible trajectory,
whereas the functional gradient of the pose term, $\nabla \mathcal{F}_{pose}$, is responsible for matching the end-effector poses $\widetilde{\mathcal{X}}$.
Even though we can update the trajectory by considering the three functional gradients simultaneously,
we found that it can lead to conflicts between each functional gradient.

To alleviate this problem, we present a two-stage gradient descent (TSGD) that
consists of  two parts of updating to make a feasible trajectory and updating the trajectory
to match closer to $\mathcal{X}$.
The TSGD is repeated in which each odd iteration updates the trajectory using $\nabla \mathcal{F}_{obs}$ and $\nabla \mathcal{F}_{smooth}$, and in which even iteration updates the trajectory using $\nabla \mathcal{F}_{pose}$:
\begin{equation}
    \xi_{i+1}=\begin{cases} 
    \xi_i - \eta_1 A^{-1} (\lambda_1 \nabla \mathcal{F}_{obs} + \lambda_2 \nabla \mathcal{F}_{smooth}), & \text{if $i$ is odd,} \\
    \xi_{i} - \eta_2 \nabla \mathcal{F}_{pose}, & \text{otherwise,}
    \end{cases}
\end{equation}
where $\eta$ is a learning rate, and $A$ is from an equally transformed representation of the smooth term, $\mathcal{F}_{smooth}=\frac{1}{2} \xi^T A \xi + \xi^T b + c$;
$A^{-1}$ acts to retain smoothness and to accelerate the optimization by having a small amount of impact on the overall trajectory~\cite{zucker2013chomp}.
Since $\mathcal{F}_{obs}$ and $\mathcal{F}_{smooth}$ have a correlation between consecutive joint configurations, it is effective for updating the trajectory with covariant gradient descent using $A^{-1}$.
On the other hand, $\mathcal{F}_{pose}$ does not consider 
consecutive joint configurations, thus we do not use $A^{-1}$.  We set TSGD to
perform $60$ iterative updates for refining a trajectory on average.

Our goal is to avoid collisions and to follow the given end-effector path in
the Cartesian workspace, while we achieve the goal through the manipulation of
joint configurations in the configuration space (C-space).
As a result, while
$\mathcal{F}_{pose}$ and $\mathcal{F}_{obs}$ are computed in workspace using
FK, $\nabla \mathcal{F}_{pose}$ and $\nabla \mathcal{F}_{obs}$ should be
defined in C-space for calculating the change of joint configurations.

\Skip{
    Thus, CHOMP defines the functional gradient as: 
    \begin{equation}
    \nabla \mathcal{F} = \frac{\partial{v}}{\partial q} - \frac{d}{dt}\frac{\partial v}{\partial q^{'}},
    \label{eq:functional_gradient}
    \end{equation}
    where $v$ indicates the formula of the objective function inside the sigma\YOON{What is the sigma here? Also, what is $q^'$?}.
}
Since $\frac{1}{2}\left\| \mathrm{x}_i - FK(q_i) \right\|^2$ of Eq.~\ref{eq:objective_pose} can be represented as $\frac{1}{2}(\mathrm{x}_i - FK(q_i))^T(\mathrm{x}_i - FK(q_i))$, 
we can derive the functional gradient of the pose term,
$\nabla \mathcal{F}_{pose}$, by the following:
\begin{equation}
\begin{aligned}
\nabla \mathcal{F}_{pose}(q_i) = J^{T}(\mathrm{x}_i-FK(q_i)),
\end{aligned}
\label{eq:functional_gradient_pose}
\end{equation}
where $J = \frac{dx}{dq} \in \mathbb{R}^{6 \times d}$ is the Jacobian matrix.

\Skip{
	Note that the Jacobian-based approach can lead to false-negative failures as
	explained in Sec.~\ref{sec:2_inverse_kinematics}.
	We, however, found that our TSGD method ameliorates the problem in practice,
	thanks to its update process that iteratively switches functional
	gradients\YOON{unclear. why does it help?}.
}


\Skip{
As $\mathcal{F}_{obs}$ takes into account the positions and velocities in workspace, CHOMP derives the functional gradient as $\nabla \mathcal{F}_{obs} = \frac{\partial{h}}{\partial q} - \frac{d}{dt}\frac{\partial h}{\partial q^{'}}$, where $h$ indicates inside the sigma of the Eq.~\ref{eq:objective_obs}.
}

$\nabla \mathcal{F}_{obs}$ can be derived as the following~\cite{zucker2013chomp}:
\begin{equation}
\nabla \mathcal{F}_{obs}(q_i) = \sum_{p}^{\mathcal{P}} J_p^T \left(\|x_{i, p}^{'}\|[(I-\hat{x}_{i, p}^{'}\hat{x}_{i, p}^{'T})\nabla c(x_{i, p}) - c(x_{i, p}) \kappa]\right), 
\end{equation}
where $x_{i, p}$ is equal to $x(q_i, p)$, 
$c_{i, p}$ to $c(x_{i, p})$, $\hat{x}_{i, p}^{'}$ is the normalized velocity vector, and $\kappa = \|x_{i, p}^{'}\|^{-2}(I-\hat{x}_{i, p}^{'}\hat{x}_{i, p}^{'T})x^{''}$ is the curvature vector. 
Since $\mathcal{F}_{smooth}$ is already represented in C-space, $\nabla \mathcal{F}_{smooth}$ is represented simply as $\nabla \mathcal{F}_{smooth}(\xi) = A\xi+b$.

Once a start configuration $q_0$ is given, we iteratively update from its next
configuration, $q_1$, to the goal configuration $q_n$, 
based on our TSGD.
In our problem, however,
the start configuration may not be given, while an end-effector path is given.
\Skip{
For example, a start configuration $q_0$ can be provided as the current configuration, or we may need to find $q_0$ as the best start configuration given a end-effector pose. 
When we do not have the start configuration,
while $q_0$ is fixed.
}
\Skip{
At that case, 
we perform to find a start configuration $q_0$ based on the quantum annealing.
}

\subsection{Iterative Exploration with Local Update}
\label{sec:3_iterative_local_search}

\Skip{
\setlength{\textfloatsep}{5pt} 
\begin{algorithm}[t] 
	\vspace{0.1cm}
	\DontPrintSemicolon 
	\KwIn {$t_{max}$: planning time}
	$t \gets 0 $\\
	\While{$t < t_{max}$} {
		$\xi \gets $ GenerateNewTrajectory() \\
	    $\xi \gets $ Two-StageGradientDescent($\xi$) \\
	    $\xi \gets $ CompareTrajectories($\xi, \xi_{new}$) \\ 
	}
	\caption{TORM} \label{alg:torm}
\end{algorithm}
}

\Skip{
\begin{figure}[t]
	\vspace{0.1cm}
	\centering 
	\subfigure [Initial trajectory]{
		\includegraphics[width=1.00in]{Fig/ils_init.pdf}
		\label{fig:ils_init} 
	}
	\subfigure 
	[Local minium (2.67e-2)]
	{ 
		\includegraphics[width=1.0in]{Fig/ils_local.pdf}
		\label{fig:ils_local} 
	} 
	\subfigure [Final trajectory (4.92e-6)]{
		\includegraphics[width=1.0in]{Fig/ils_final.pdf}
		\label{fig:ils_final} 
	}
	\caption{
		Red lines represent the given end-effector path, and blue lines
		are computed end-effector paths during our iterative process.
		Numbers within parenthesis indicate the pose error
		(Sec.~\ref{sec:4_evaluation}).  \YOON{need to be adjusted}We
		repeatedly find a new
		initial trajectory and refine the trajectory using our local
		optimization.  Even if the trajectory falls into a local
		minimum, we get out of it through iterative exploration and
		robustly minimize the pose error.		
	} 
	\label{fig:ils}
	\vspace{0.1cm}
\end{figure}
}

We can optimize a trajectory based on our update rule
(Sec.~\ref{sec:3_optimization_process}), but there is a probability that
the updated trajectory is sub-optimal due to the local natures of our update
method.  Furthermore, there can be many surrounding local minima in 
our solution space, since we incorporate three different properties of constraints into the objective function.

To avoid getting stuck in local minima, we perform iterative exploration with
local updates. Note that a redundant manipulator can have multiple joint
configurations at a single pose, thus we can construct many candidate trajectories.
By utilizing this property, we generate different new trajectories, which are locally refined based on our two-stage gradient descent
(Sec.~\ref{sec:3_optimization_process}).

Our algorithm iteratively performs two parts:
exploring a new trajectory, $\xi_{new}$, 
and locally refining the trajectory.
When a new trajectory $\xi_{new}$ that is also locally refined
has a lower cost to the existing
one, we use it to the current trajectory; if
the new trajectory violates the constraints, we reject it and thus do not use it as the current trajectory
(Sec.~\ref{sec:3_trajectory_constraints}).  The update part locally refines the 
trajectory using our two-stage gradient descent
(Sec.~\ref{sec:3_optimization_process}).  This process continues until
satisfying a given condition, e.g., running time or cost.

\begin{figure}[t]
	\vspace{0.25cm}
	\centering
	\includegraphics[width=2.5in]{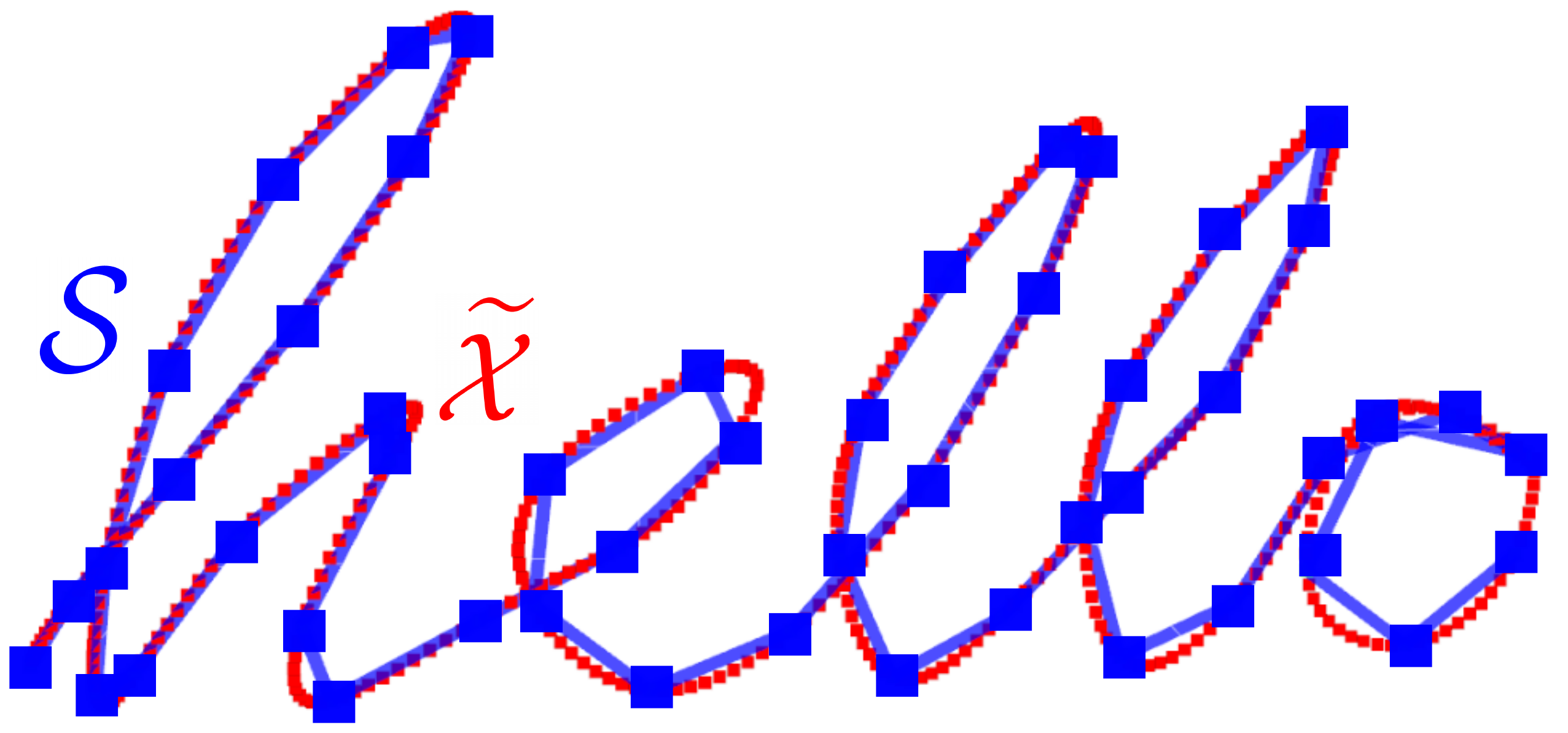}
	\caption{
		This figure shows sub-sampled poses $\mathcal{S}$
		(blue dots) in the problem of writing ``hello".
		The sub-sampled poses are extracted uniformly from finely
		divided end-effector poses $\widetilde{\mathcal{X}}$ (red
		dots).  From $\mathcal{S}$, we construct a new trajectory
		$\xi_{new}$ that starts with random IK solutions and is found
in a greedy manner minimizing our objective function; the blue lines show the	end-effector path for $\xi_{new}$.
	}
	\label{fig:simplified_poses}
	\vspace{-0.0cm}
\end{figure}

\noindent
\textbf{Exploring and generating a new trajectory.}
Each exploration part generates a new trajectory $\xi_{new}$.  We strive for finding a
potentially good trajectory for our local optimization, which considers our
objectives with smoothness, avoiding obstacles, and following a given path.
Because merely connecting start and goal configuration
can result in a sub-optimal
solution, especially in cases of complex scenarios (e.g.,
Fig.~\ref{fig:simplified_poses} and environments with external
obstacles)~\cite{holladay2016distance}.


\Skip{
Constructing a good initial trajectory faces a main challenge of considering
our objectives with smoothness, avoiding obstacles, and following a given path.
Note that there can be multiple or even an infinite number of joint
configurations at an end-effector pose.  Therefore, simply obtaining joint
configurations using IK at each end-effector pose and linearly interpolating
some of them is not effective for computing a trajectory.
}

Overall, we consider random configurations and choose one that minimizes our
objective function in a greedy manner for generating an initial trajectory. 
As the first step, 
we extract sub-sampled poses, $\mathcal{S}$, at $m$ intervals
from the end-effector poses $\widetilde{\mathcal{X}}$, since working with more
poses tends to increase the complexity of generating a trajectory
(Fig.~\ref{fig:simplified_poses}); we set $m$ to $10$. 
In the next step, we find suitable joint configurations at the sub-sampled
poses. For the first sub-sampled pose as the start end-effector pose
$\mathrm{x}_0$, we simply compute a random IK solution at the pose,  if a start
configuration is not given.
For its next pose, we compute $k$ random IK solutions at the pose and greedily
select one that minimizes our objective function when connecting it with the
configuration of the previous pose; we set $k$ to $150$.  Lastly, we connect
chosen joint configurations through linear interpolation for generating a new
trajectory, which is then locally refined by our two-stage gradient descent.

	\subsection{Trajectory Constraints}
\label{sec:3_trajectory_constraints}

During the optimization process, we may find a low cost solution, but it can
violate several constraints.
For example, a trajectory can have collisions with obstacles or
self-collisions, even though the trajectory accurately follows the given
end-effector pose.  Hence, we check collisions every time we find a better
trajectory during our iterative exploration.

In addition to the collision checking, a manipulator commonly has several constraints that are satisfying lower and upper limits of joints,
velocity limits, and kinematic singularities for joints.
In the case of lower and upper limits of joints, it is traditionally handled by performing $L_1$ projection that resets the violating joints value to its limit value. 
To retain smoothness, we use a smooth projection used by
CHOMP~\cite{zucker2013chomp} during the update process. The smooth projection
uses the Riemannian metric $A^{-1}$.
In other words, an updated trajectory, $\tilde{\xi}$, is
defined as $\tilde{\xi} = \xi + \alpha A^{-1} v$, where $v$ is the vector of
joint values, and $\alpha$ is a
scale constant. This process is repeated until there is no violation.

For other constraints like the velocity limit, we check them together while checking collisions. 
The velocity limit for joints is checked by computing the velocities of joints between $q_{i-1}$ and $q_{i}$ for a given time interval, $\Delta t$.
Another constraint is the kinematic singularity that is a point where the robot is unstable, and it can occur when the Jacobian matrix loses full rank.
To check kinematic singularities, we use the manipulability measure~\cite{yoshikawa1985manipulability} used by RelaxedIK~\cite{rakita2018relaxedik}. At a high level, it avoids making the manipulability measure less than a certain value that is computed by random samples.
Note that our method returns the lowest cost trajectory guaranteed through checking constraints for constructing
a feasible trajectory.
	\begin{figure}[t]
	\vspace{0.1cm}
	\centering 
	\subfigure [EIGS (5.43e-3)]{
		\includegraphics[width=1.57in]{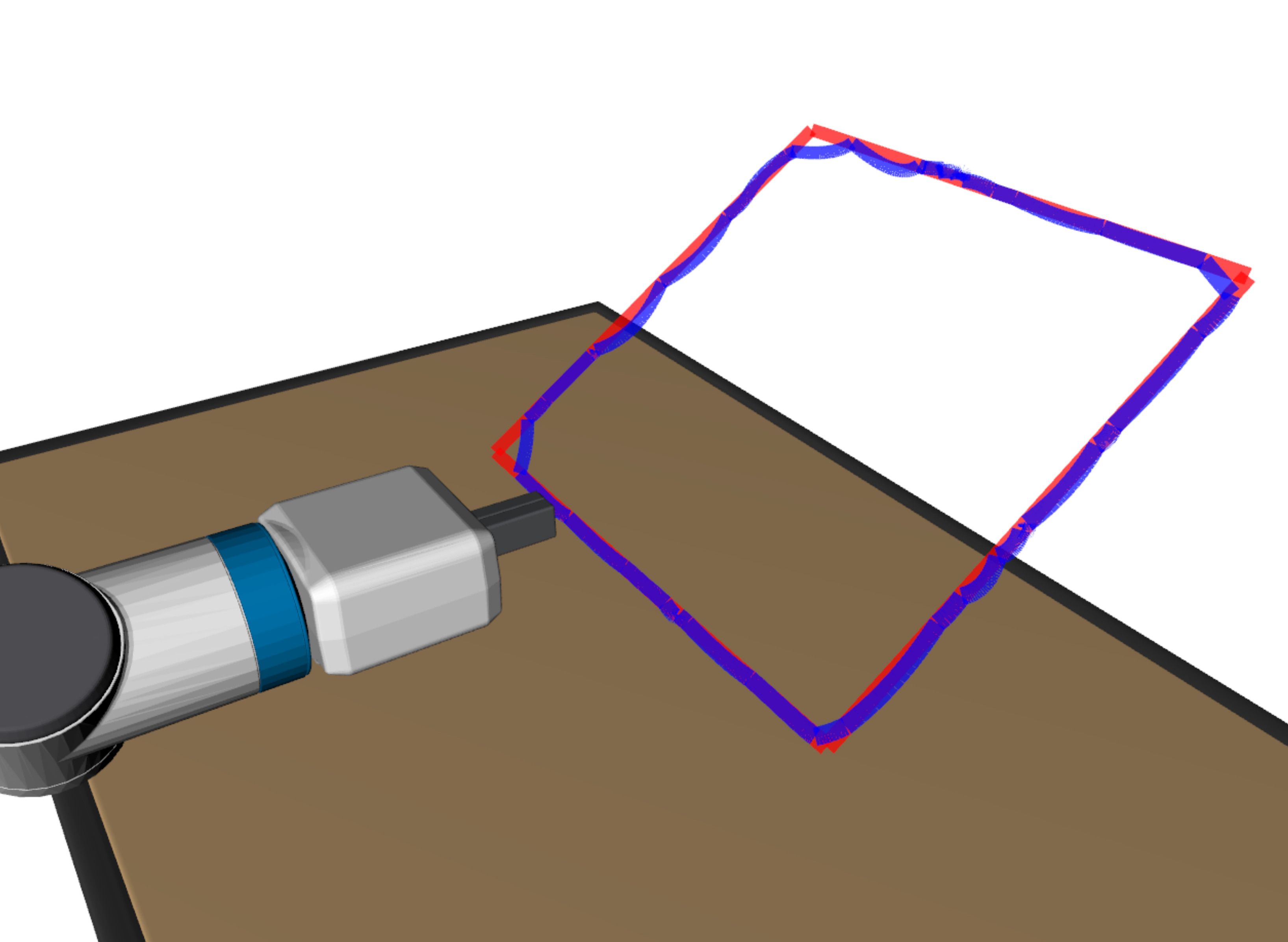}
		\label{fig:traj_square_eigs} 
	}
	\subfigure 
	[Our method (1.92e-5)]
	{ 
		\includegraphics[width=1.568in]{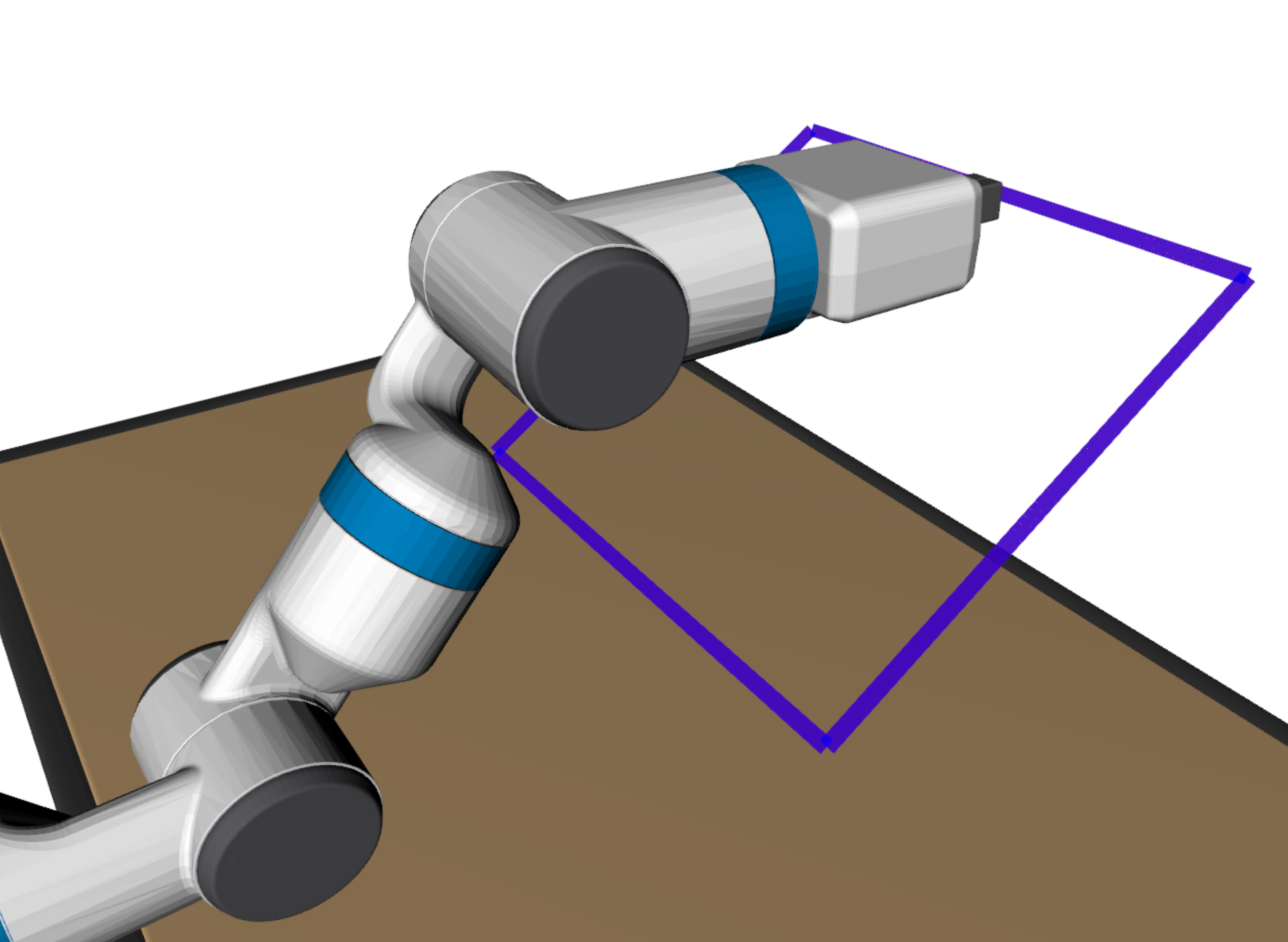}
		\label{fig:traj_square_ours} 
	} 
	\subfigure [EIGS (6.63e-3)]{
		\includegraphics[width=1.57in]{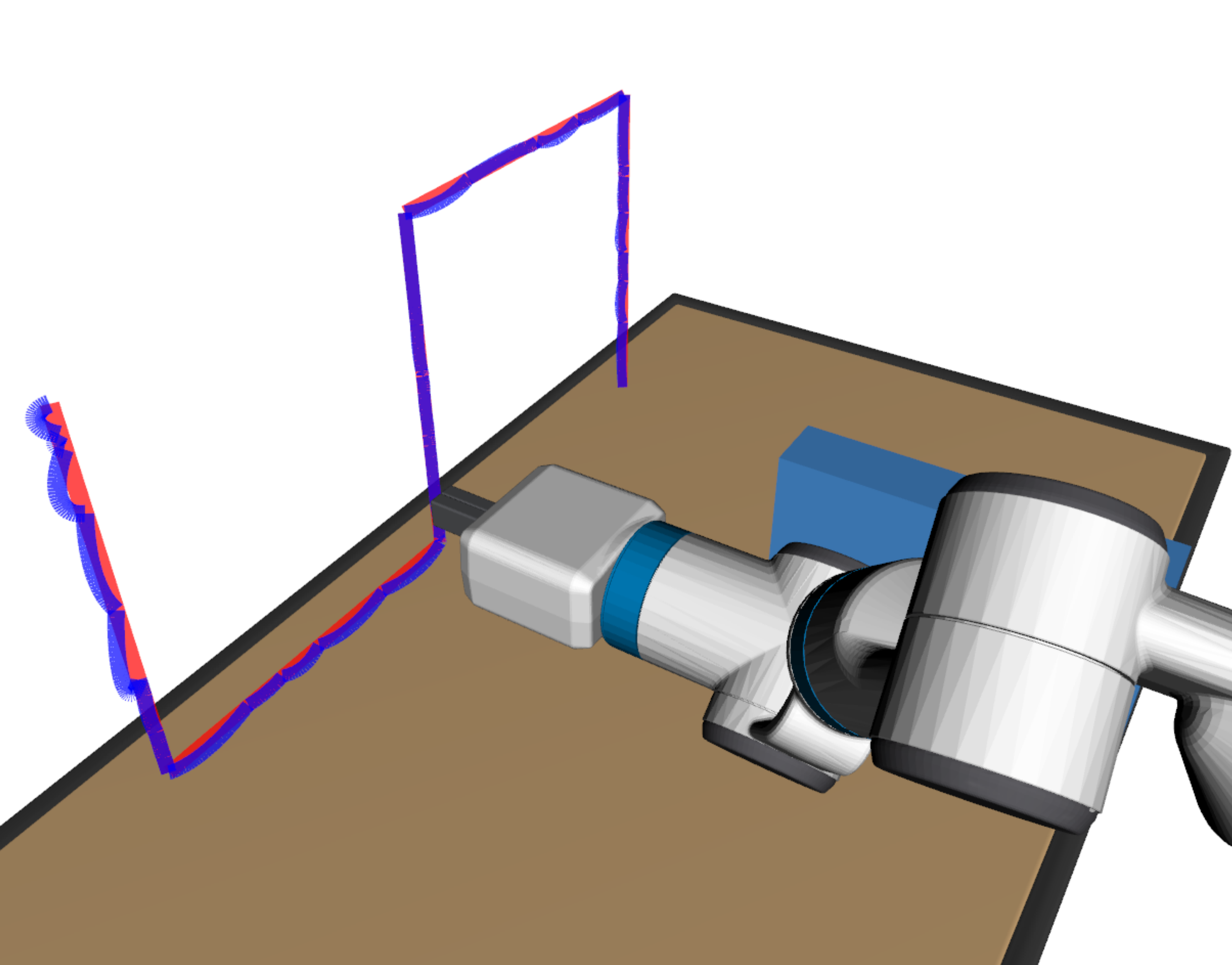}
		\label{fig:traj_s_eigs} 
	}
	\subfigure 
	[Our method (3.16e-5)]
	{ 
		\includegraphics[width=1.568in]{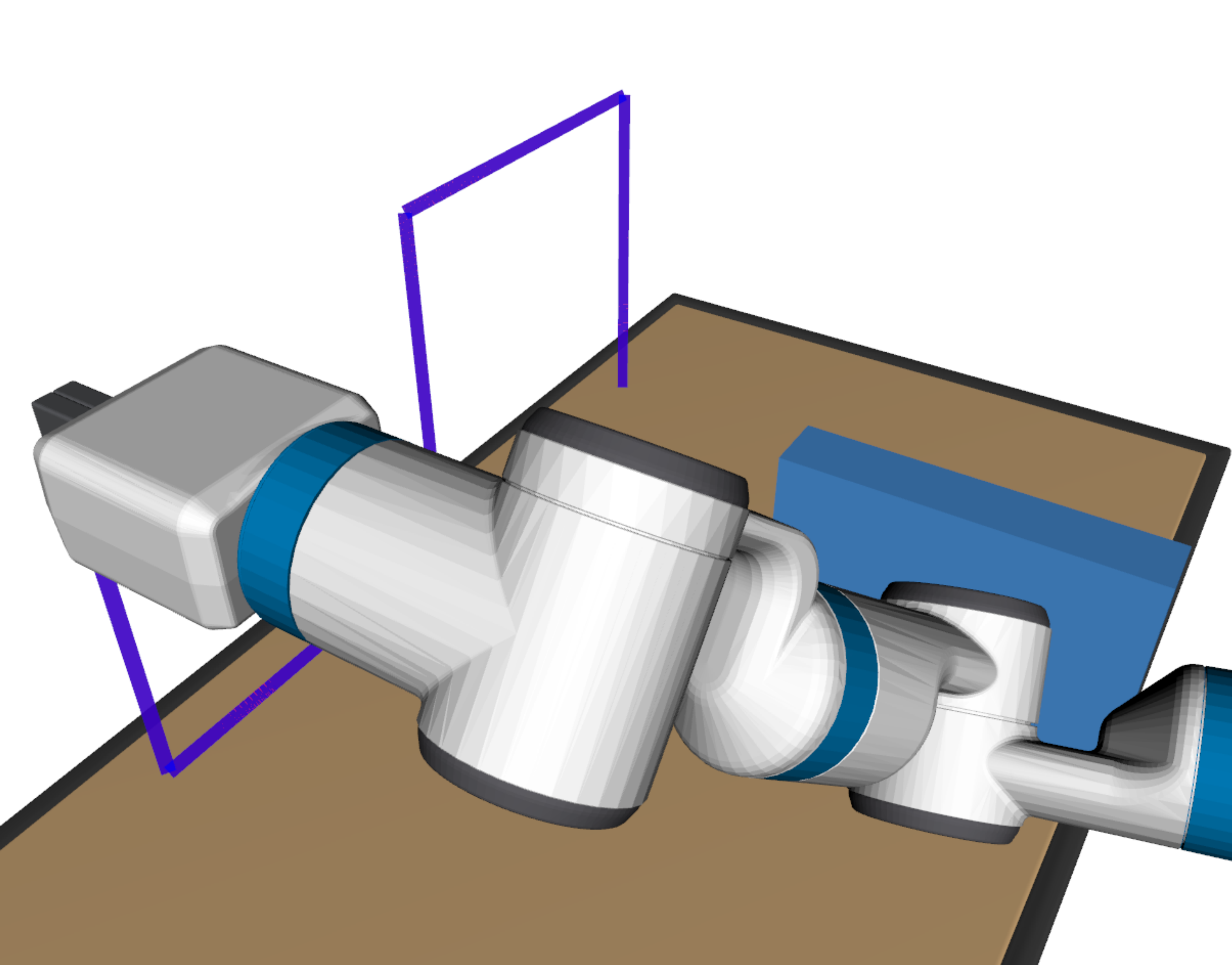}
		\label{fig:traj_s_ours} 
	} 
	\caption{
		This shows the visualization results, where red lines are the
		given paths, and blue lines are computed end-effector paths.
		Numbers within parenthesis indicate the pose error.  $(a)$ and
		$(b)$ are to trace the square, and $(c)$ and $(d)$ are to trace
		the ``S".  
		We must avoid the table across all scenes, and additionally
		consider the blue box in $(c)$ and $(d)$.
		Results of EIGS, $(a)$ and $(c)$, show noticeable pose errors
		in several parts.  Our methods shown in $(b)$ and $(d)$
		accurately follow the given path with smaller numerical errors.
	} 
	\label{fig:obs_problems}
	\vspace{-0.4cm}
\end{figure}
\begin{figure}[t]
	\vspace{0.1cm}
	\centering 
	\subfigure [RelaxedIK (5.39e-2)]{
		\includegraphics[width=1.57in]{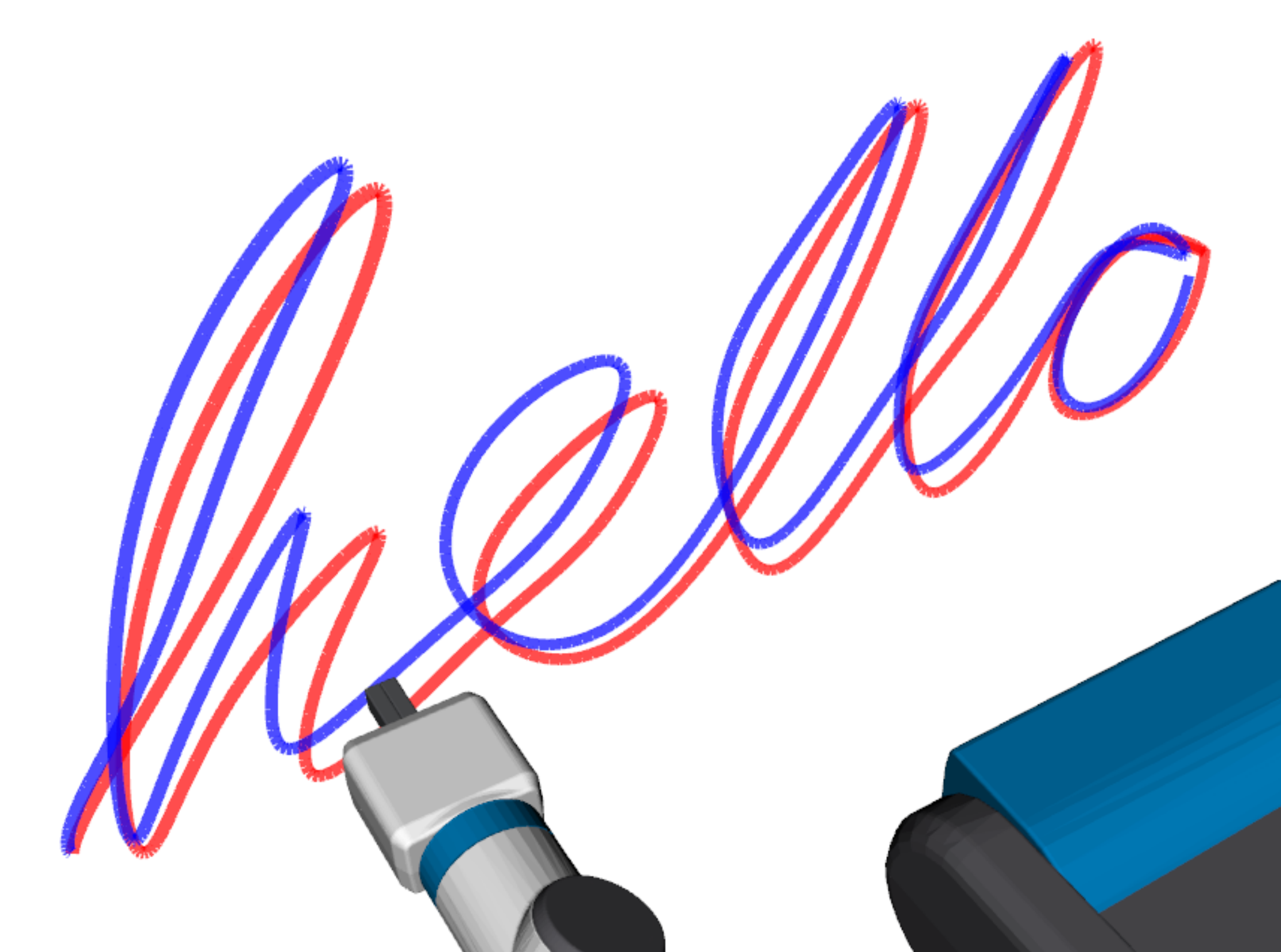}
		\label{fig:traj_hello_relaxed} 
	}
	\subfigure 
	[Stampede (5.08e-5)]
	{ 
		\includegraphics[width=1.568in]{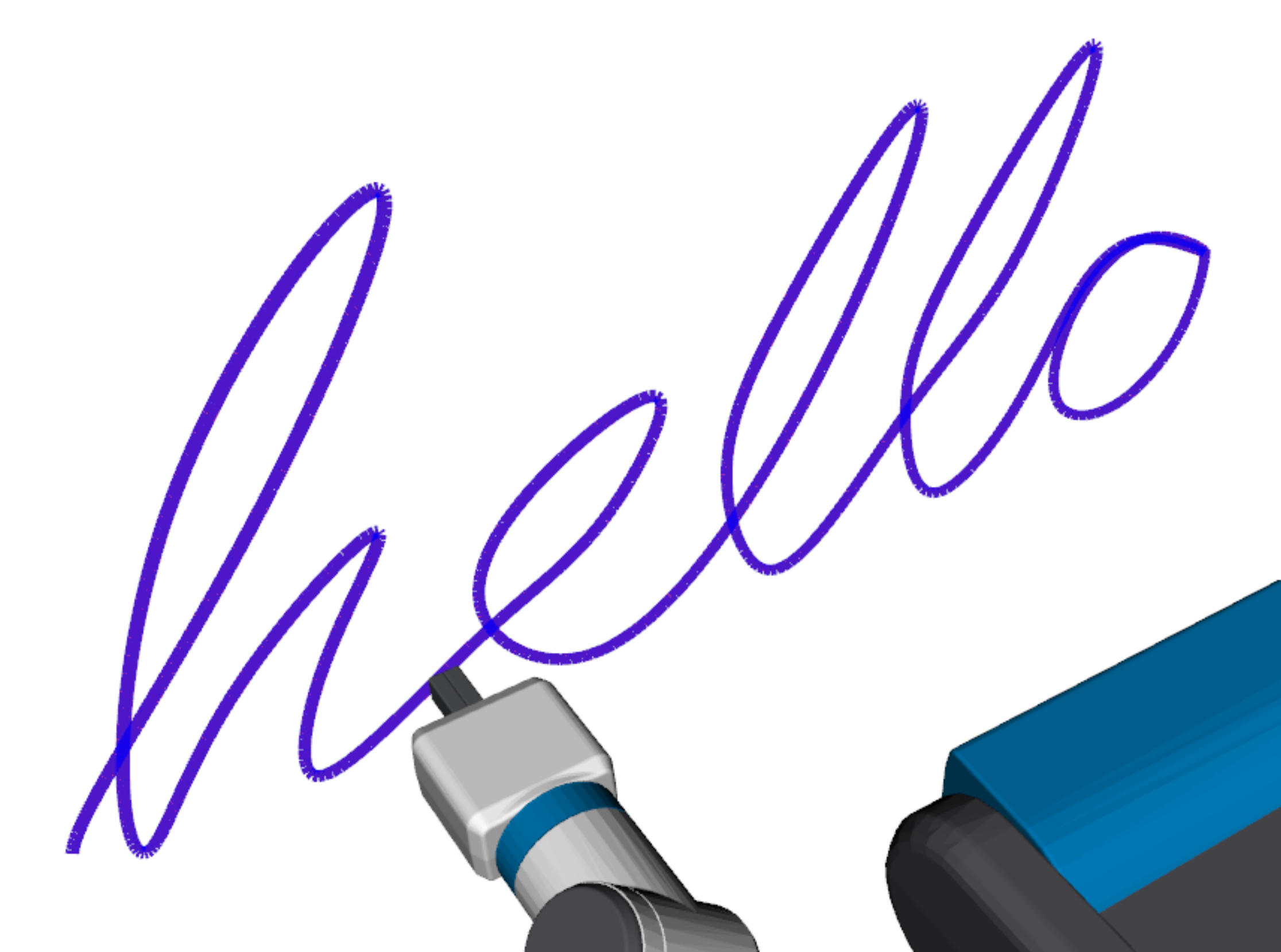}
		\label{fig:traj_hello_stampede} 
	} 
	\subfigure [EIGS (1.92e-2)]{
		\includegraphics[width=1.57in]{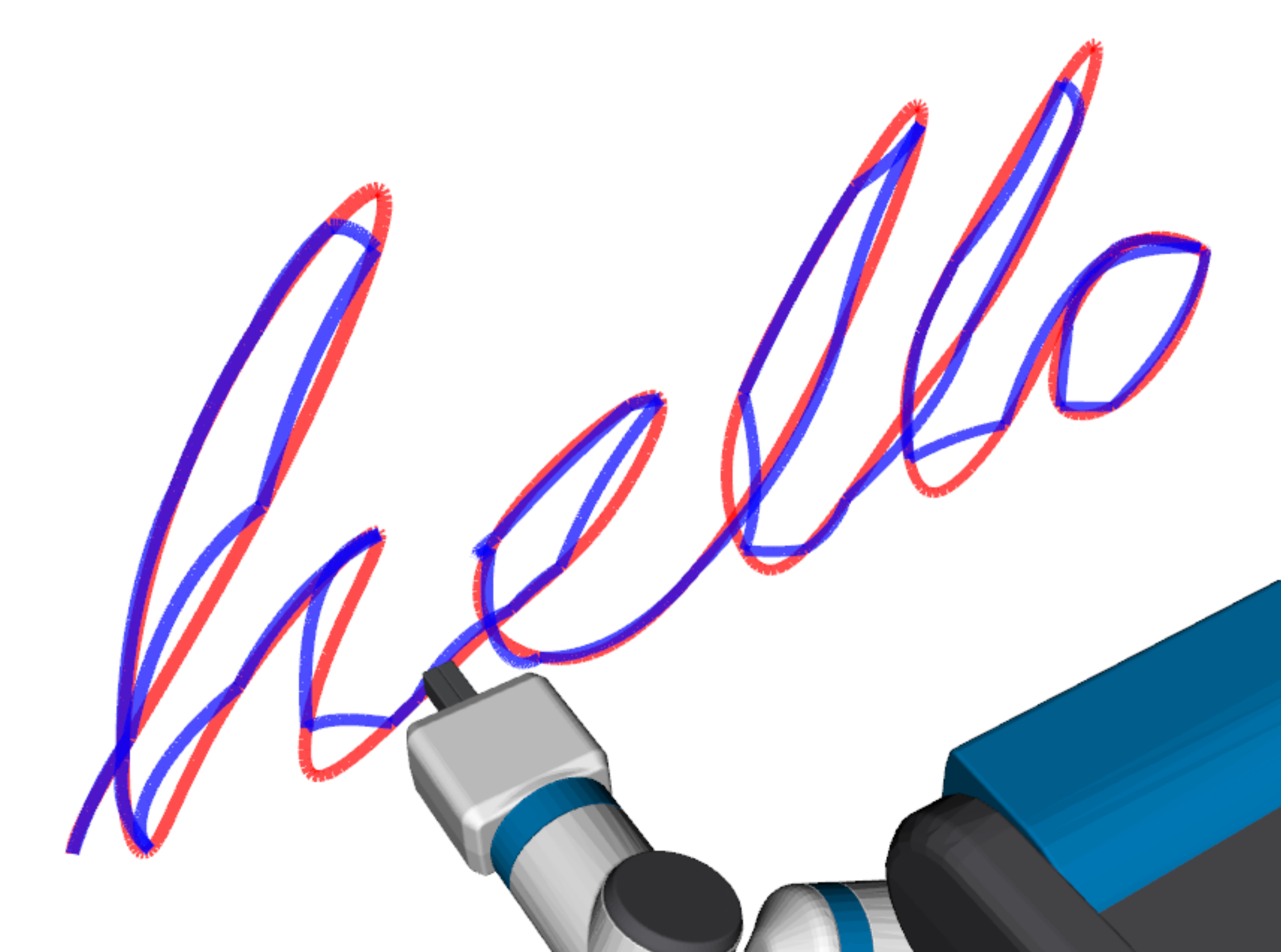}
		\label{fig:traj_hello_eigs} 
	}
	\subfigure 
	[Our method (5.03e-5)]
	{ 
		\includegraphics[width=1.568in]{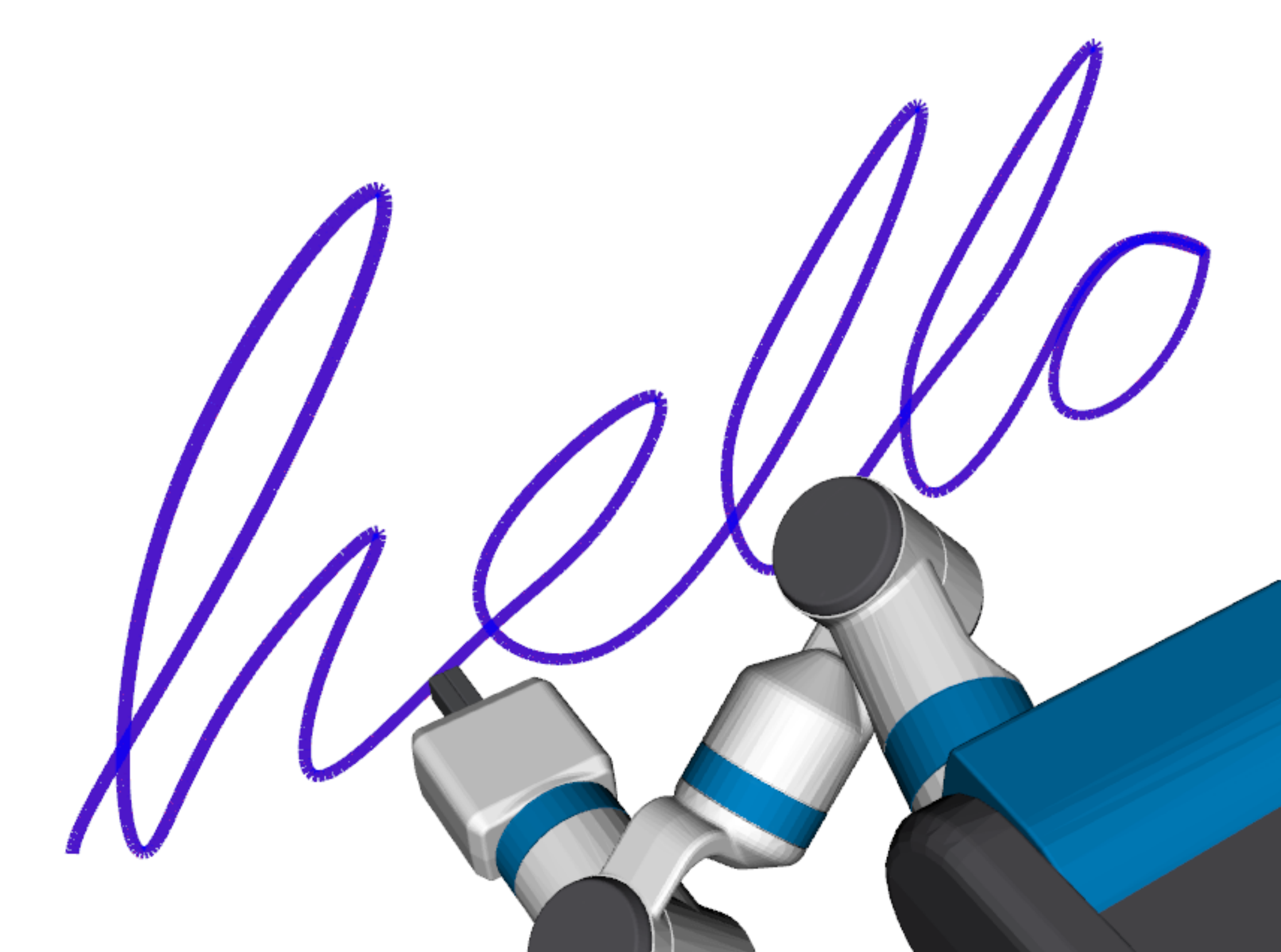}
		\label{fig:traj_hello_ours} 
	} 
	\caption{
		This shows the visualized results of writing ``hello" w/o obstacles for different methods.
		Red lines are the given paths, and blue lines are computed end-effector paths. Numbers within parenthesis indicate the pose error.
		$(a)$ and $(c)$ show relatively high pose errors, while $(b)$
		Stampede and $(d)$ ours show accurately following the given path.
	} 
	\label{fig:results_hello}
	\vspace{-0.4cm}
\end{figure}

\section{Experiments and Analysis}
\label{sec:4}

We use the Fetch robot, whose manipulator arm has 7-DoF for various tests.
We report the average performance by performing $20$ tests with 
a machine that has 3.4 GHz Intel i7-6700 CPU and 16 GB RAM.
Also, we set $\lambda_1$ = $1.8$, $\lambda_2$ = $\frac{5.0}{n+1}$, $\eta_1$ = $0.03$ and $\eta_2$ = $1.0$, where $n+1$ is the number of the end-effector poses $\widetilde{\mathcal{X}}$. 

In this setting, the two-stage gradient descent, generating new trajectory, and checking constraints 
of our method take 67\%, 32\%, and 1\% of the overall running time;
the two-stage gradient descent is frequently iterated to refine
the trajectory, as the main update operation.

In this experiment, we compute the distance between end-effector poses using
a weight sum of the Euclidean distances of the translational and rotational
parts, which was used in the work of Holladay et
al.~\cite{holladay2019minimizing}; the weight of the rotational distance over
the translational distance is $0.17$.



\Skip{
	\YOON{What are issues depending on values of n?}
	\MC{In the same path length, larger $n$ increases accuracy but increases
		computational time.  If $n$ is too small and the distance between poses is too
		far, it is difficult to find feasible configurations between two adjacent poses
		due to velocity limits of joints.  Stampede assumes that an accurate result is
		practically obtained if two adjacent poses are short enough.  }
}

Results with the real Fetch robot is shown in the accompanying video.

\subsection{Experiment setting}
\label{sec:4_exp_setting}

We prepare two problems with external obstacles and two non-obstacle problems
to evaluate and compare our method with state-of-the-art methods,
RelaxedIK~\cite{rakita2018relaxedik}, Stampede~\cite{rakita2019stampede}, and
the work of Holladay et al.~\cite{holladay2019minimizing}. 
In this section, we call the work of Holladay et
al.~\cite{holladay2019minimizing} EIGS taken from their paper title.

Two problems with obstacles (Fig.~\ref{fig:obs_problems}) are the square
tracing with the table and the ``S" tracing with the table and blue box.  At
these problems, we do not test for RelaxedIK and Stampede, since these prior
methods did not consider external obstacles.
To compare ours against those prior methods, we prepare two non-obstacle
problems, rotating $\pm 45$ degrees in the direction of pitch and yaw, and
writing ``hello" (Fig.~\ref{fig:results_hello}), by adapting problems used in those methods; we just change writing ``icra" to ``hello".

In two problems including external obstacles, we fix the start configuration
located close to the obstacles, as shown the Fig.~\ref{fig:traj_square_ours}
and Fig.~\ref{fig:traj_s_ours}.  On the other hand, we do not fix the start
configuration for non-obstacle problems.

We used the code of RelaxedIK and Stampede that are provided by authors
through their websites.  For RelaxedIK, Stampede, and our method, the
end-effector paths of all problems are finely divided and are fed into all the
tested methods; in our experiment, the distance between two divided
end-effector poses is about $0.005$, following the 
protocol adopted in \cite{rakita2019stampede}.
\Skip{
	\MC{I don't know that the paragraph below is needed.}
	\MC{
		In Stampede, they assume that if the interval between two adjacent
		poses is short enough, a feasible and accurate trajectory can be
		practically achieved~\cite{rakita2019stampede}.
		We follow their assumption, since they showed the highly-accurate results. 
	}
}
On the other hand, EIGS initially splits the end-effector path and gradually
breaks down the path during the planning.

\Skip{
	Specifically, we split the end-effector path into $10$ initially, and samples
	$30$ IK solutions at each end-effector pose at a time. 
	These parameters were We experimentally found these parameters 
	to
	experimentally reasonable values considering their anytime property, which
	quickly finds an initial trajectory and refine the
	trajectory~\cite{karaman2011anytime}.
}

\begin{figure}[t]
	\vspace{0.1cm}
	\centering 
	\subfigure [Result of square tracing w/ obs.] {
		\includegraphics[width=1.58in]{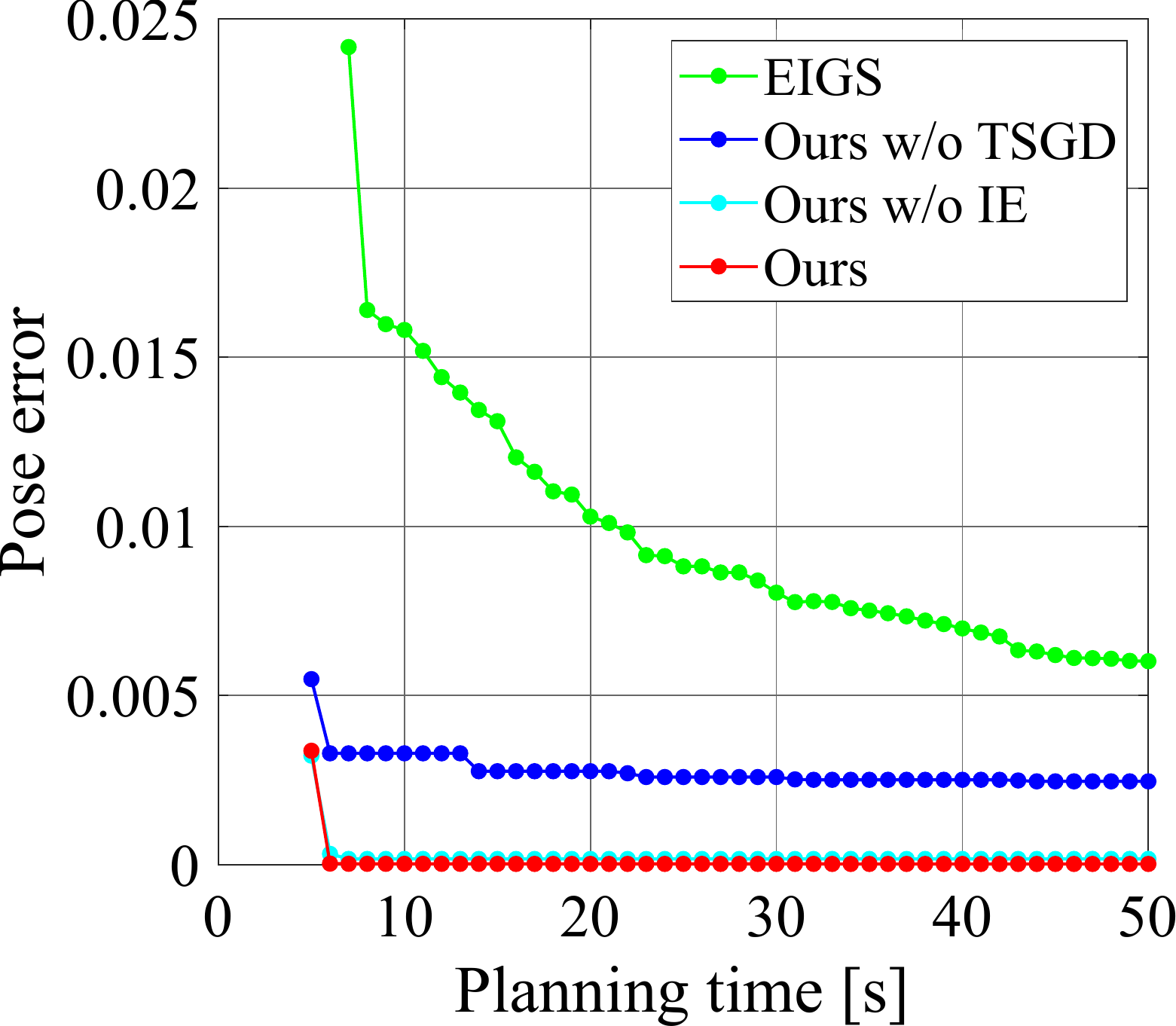}
		\label{fig:res_square_obs} 
	} 
	\subfigure [Result of ``S" tracing w/ obs.] { 
		\includegraphics[width=1.58in]{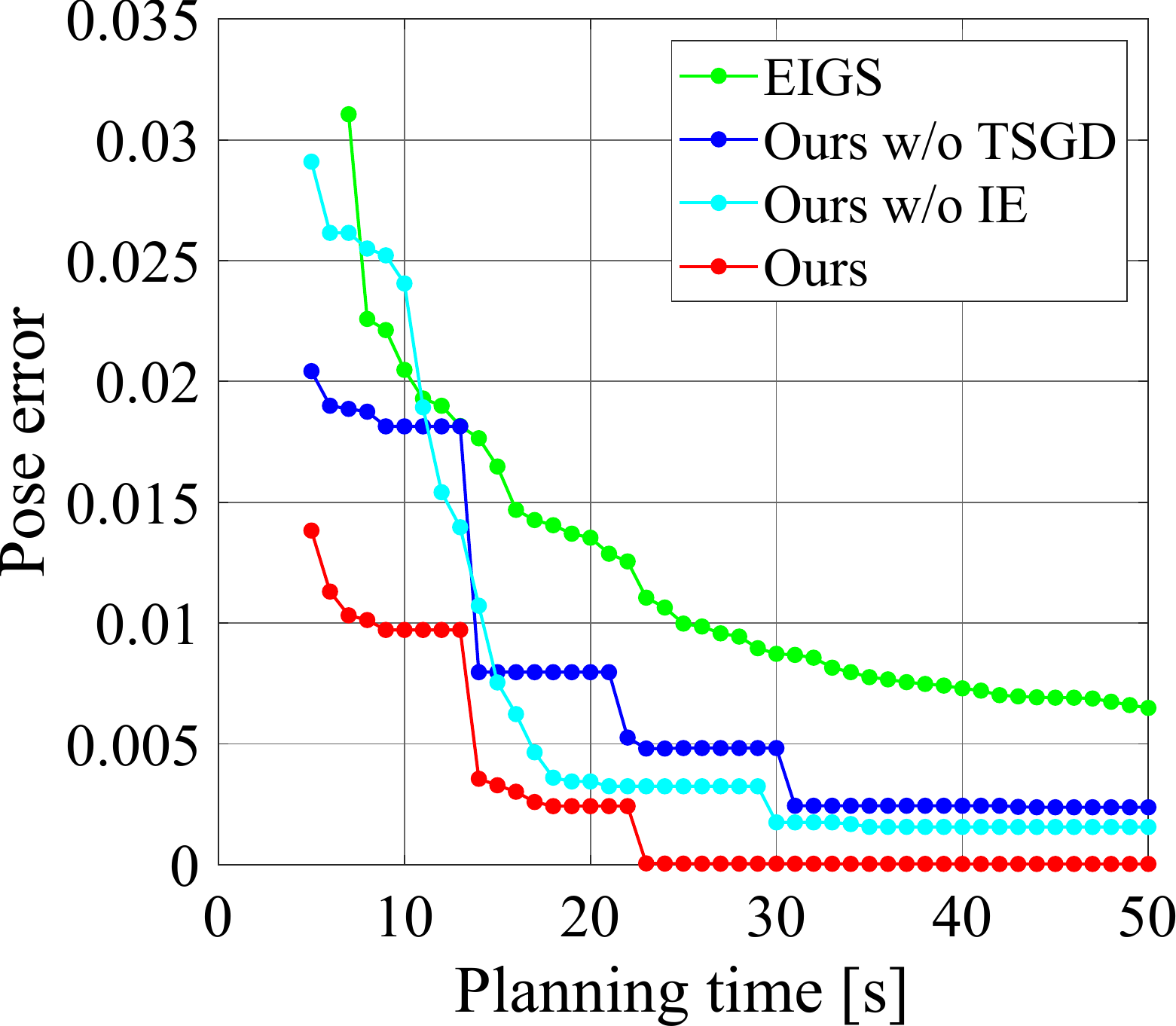}
		\label{fig:res_s_obs} 
	}
	\subfigure [Result of rotation task w/o obs.] {
		\includegraphics[width=1.58in]{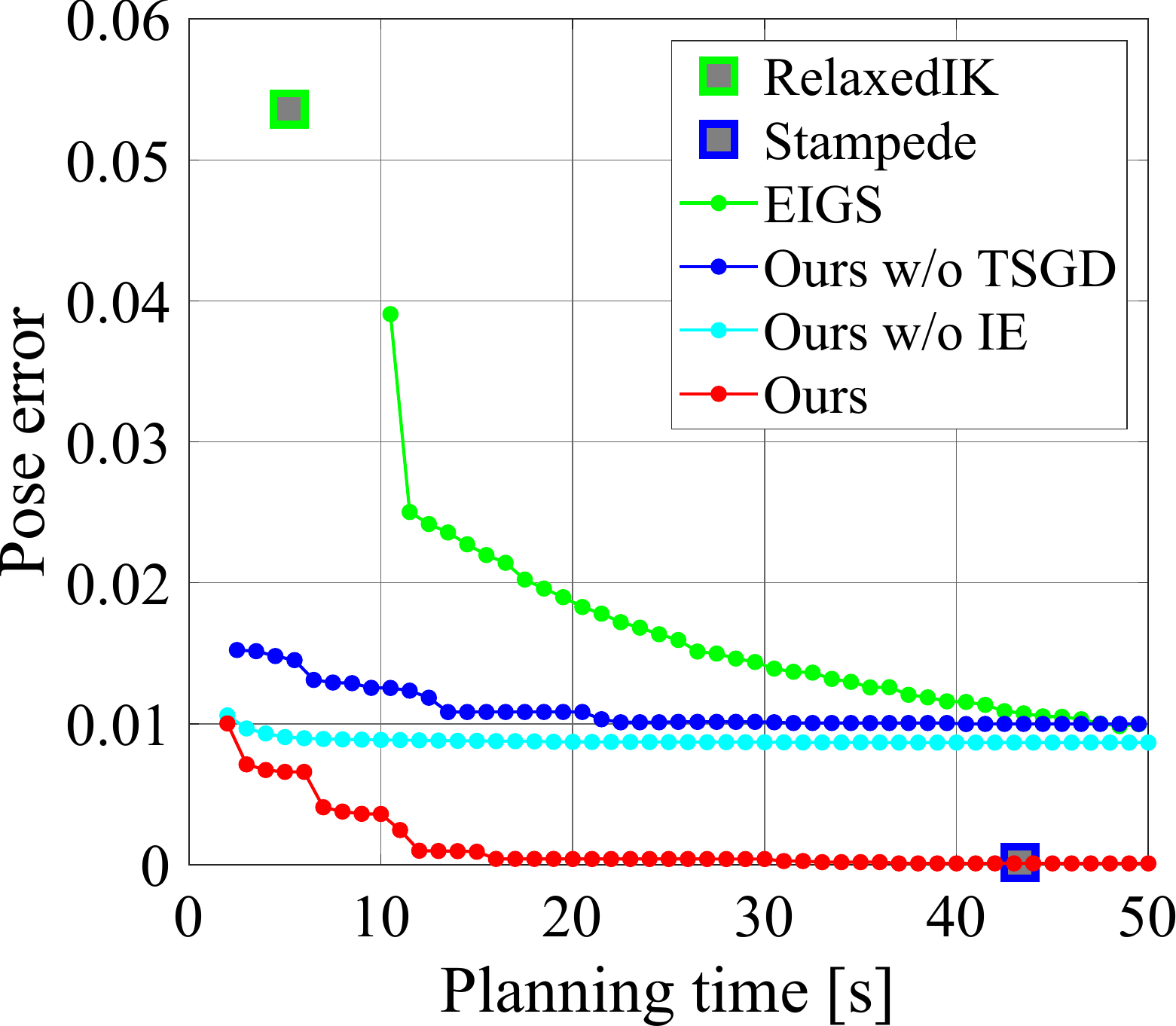}
		\label{fig:res_rotation} 
	} 
	\subfigure [Result of writing ``hello" w/o obs.] { 
		\includegraphics[width=1.58in]{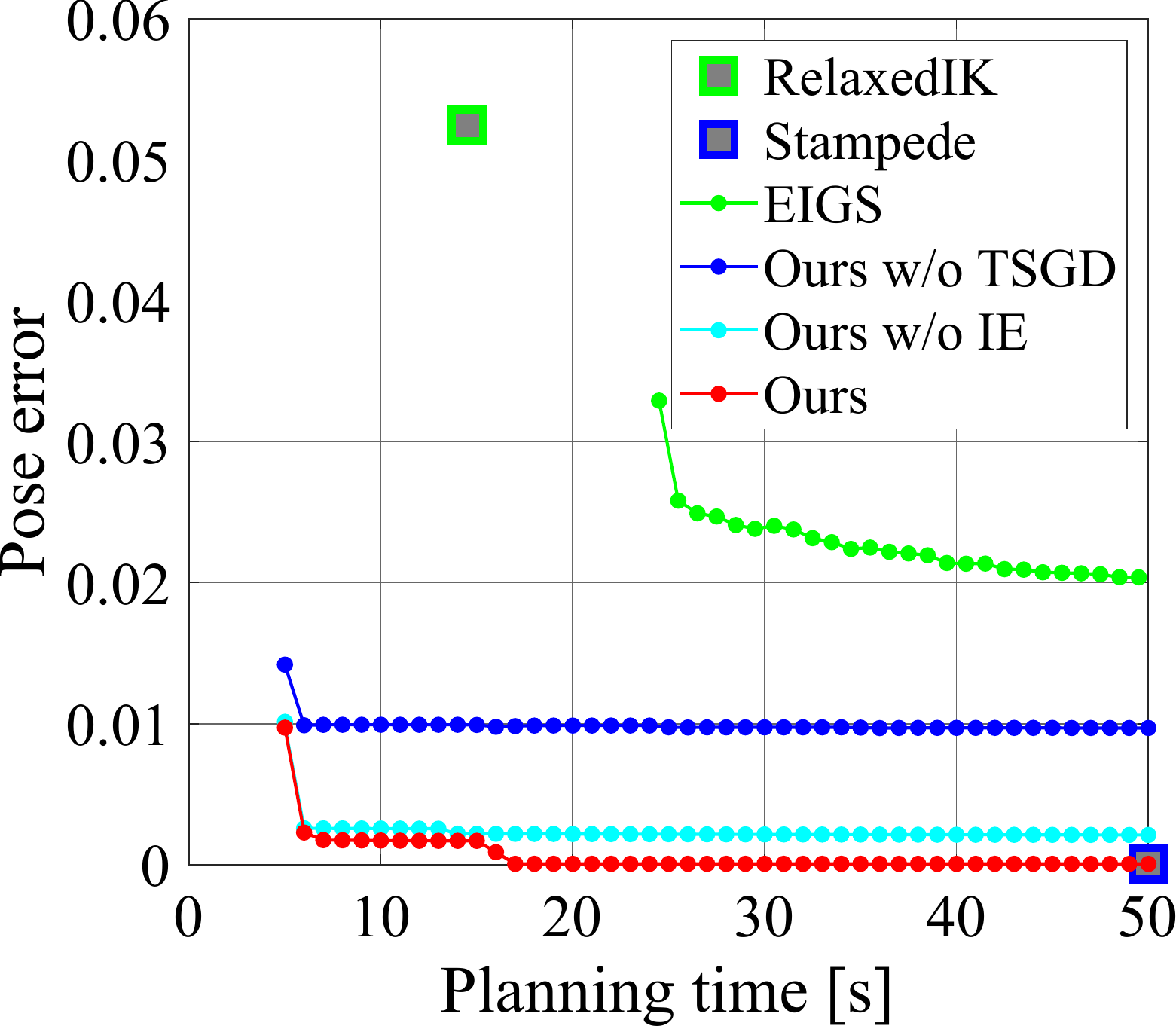}
		\label{fig:res_hello} 
	} 
	\caption{
		This shows the pose error over the planning time of state-of-the-art methods and our method, including ablated methods that are our method w/o the two-stage gradient descent (TSGD) and w/o the iterative exploration (IE). 
		Since $(a)$ and $(b)$ are the results in the environment with external obstacles, RelaxedIK and Stampede are excluded from the experiments.
		We visualize graphs once an initial solution is computed.
	} 
	\label{fig:results_chart}
	\vspace{-0.2cm}
\end{figure}

\begin{table}[t]
	\vspace{0.25cm}
	\centering
	\scriptsize
	\renewcommand \arraystretch{1.4}
	\setlength{\tabcolsep}{1.8pt}
	\caption{ 
		Results of different methods, EIGS, RelaxedIK, Stampede, and
		ours given an equal planning time. RelaxedIk is a real-time
		planner and does not provide better trajectories with more
		planning time; we provide its results while it cannot be
		compared in the equal-time comparison.
		EIGS and ours consider external
		obstacles, while RelaxedIK and Stampede only check
		self-collision.
	}
	\begin{tabular}{|c|c|c|c|c|c|c|c|c|}
		\hline
		\multicolumn{2}{|c|}{~} & \multicolumn{3}{c|}{Pose error} & ~\multirow{2}{*}{TL}~ & ~\multirow{2}{*}{IST}~ & ~\multirow{2}{*}{PT}~ \\
		\cline{3-5}
		\multicolumn{2}{|c|}{~} & Average & Min. & Max. & ~ & ~ & ~ \\
		\hline
		\hline
		~Square tracing~ & EIGS & 5.99e-3 & 3.92e-3 & 8.68e-3 & 23.2 & 7.1 & ~\multirow{2}{*}{50}~ \\
		\cline{2-7}
		w/ obs. (n = 304) & Ours & \textbf{~1.91e-5~} & \textbf{~1.01e-5~} & \textbf{~7.65e-5~} & \textbf{9.5} & \textbf{5.2} & ~ \\
		\hline
		\hline
		``S" tracing & EIGS & 6.50e-3 & 4.56e-3 & 1.06e-2 & 20.5 & 7.0 & ~\multirow{2}{*}{50}~ \\
		\cline{2-7}
		w/ obs. (n = 300) & Ours & \textbf{2.40e-5} & \textbf{6.08e-6} & \textbf{8.50e-5} & \textbf{7.8} & \textbf{5.2} & ~ \\
		\hline
		\hline
		~\multirow{2}{*}{Rotation task}~ & ~RelaxedIK~ & 5.36e-2 & 4.13e-2 & 1.90e-1 & ~\textbf{11.1}~ & - & \textbf{5.2} \\
		\cline{2-8}
		~\multirow{2}{*}{w/o obs.}~ & Stampede & 9.42e-5 & 8.72e-5 & \textbf{9.95e-5} & 12.2 & 43 & ~ \\
		\cline{2-7}
		~\multirow{2}{*}{(n = 208)}~ & EIGS & 1.07e-2 & 7.53e-3 & 1.82e-2 & 16.0 & 11 & 43 \\
		\cline{2-7}
		~ & Ours & \textbf{7.99e-5} & \textbf{6.83e-5} & 1.21e-4 & 13.8 & \textbf{1.8} & ~ \\
		\hline
		\hline
		~\multirow{2}{*}{Writing ``hello"}~ & RelaxedIK & 5.25e-2 & 4.86e-2 & 5.99e-2 & \textbf{22.1} & - & \textbf{15} \\
		\cline{2-8}
		~\multirow{2}{*}{w/o obs.}~ & Stampede & 5.21e-5 & 4.93e-5 & \textbf{6.03e-5} & 24.2 & 50 & ~ \\
		\cline{2-7}
		~\multirow{2}{*}{(n = 496)}~ & EIGS & 2.02e-2 & 1.71e-2 & 2.36e-2 & 35.2 & 24 & 50 \\
		\cline{2-7}
		~ & Ours & \textbf{5.11e-5} & \textbf{4.77e-5} & 6.38e-5 & 26.4 & \textbf{5.2} & ~ \\
		\hline
	\end{tabular}
	\begin{tablenotes}
		\item[1] TL: Trajectory length (rad). IST: Initial solution time (s). PT: Planning time (s).
	\end{tablenotes}
	\vspace{0.17cm}
	\label{tab:result}
\end{table}

\subsection{Evaluation}
\label{sec:4_evaluation}

We mainly evaluate whether the synthesized trajectory accurately follows the
given end-effector path.  Note that reported results were extracted from
feasible trajectories satisfying the given constraints
(Sec.~\ref{sec:3_trajectory_constraints}).

Fig.~\ref{fig:results_chart} and Table~\ref{tab:result} show the overall
results for different methods. 
Fig.~\ref{fig:results_chart} shows the trajectory quality of different methods, as we have more planning time up to the maximum planning budget of $50$ seconds.
On the other hand, Table~\ref{tab:result} shows a snaptshot result of the
trajetory computed at the maximum planing time budget.
For the rotation task with obstacles, we report results by $43$ seconds, which
is the initial solution time of Stampede, instead of $50$ seconds; within $50$ seconds, Stampede computed only a single trajectory at $43$ seconds.
Note that in the case of RelaxedIK, a real-time planner, it quickly synthesizes
one trajectory at one execution, but its computed trajectory tends to be
low-quality.

\Skip{
	In case of joint jerk, RelaxedIK, Stampede and our method are computed between joint configurations at finely tessellated poses.
	On the other hand, EIGS can be different intervals between end-effector poses.
	Thus, we measure the joint jerk of EIGS by divding the trajectory by joint distance, $0.05$.
	It is not fair comparsion between EIGS and others, ~~~
}

\Skip{
	Interestingly, the pose error of EIGS (green line) goes up and down slightly in
	some sections \YOON{where? I cannot see it}.
	Since EIGS uses the discrete Fr\'{e}chet distance as the approximate similarity measure of two curves, the Fr\'{e}chet distance is lowered, but the pose error can be higher \YOON{can you improve this sentence? can you see what could be problems here?}.
}

\noindent
\textbf{Comparison with state-of-the-art methods.}
Fig.~\ref{fig:results_chart} shows how different methods compute a trajectory.
EIGS and our method are improving its quality as we have more planning time, since they have the anytime property.

EIGS refines a trajectory by progressively sampling new waypoints and IK solutions. 
Nonetheless, our method improves the quality of the trajectory over EIGS, as we
have more planning time (Fig.~\ref{fig:results_chart}).  
This improvement is mainly because our method incorporates the IK solving
method into the optimization process (Eq.~\ref{eq:functional_gradient_pose}),
instead of simply using IK solutions.

In Table~\ref{tab:result}, EIGS shows the longest length of the computed
trajectory on average for all problems; we measure the length of a trajectory
using the Euclidean distance.  EIGS does not consider the distance in C-space,
since it only check the similarity measure of two curves using the discrete
Fr\'{e}chet
distance~\cite{holladay2019minimizing}.
On the other hand, other methods take into account the smoothness between joint
configurations and thus generate shorter trajectories than EIGS.


Stampede is also a sampling-based approach like EIGS, but it generates a
highly-accurate solution on average (Table~\ref{tab:result}).  Stampede
does not deal with external obstacles, but uses a neural network to check
self-collision quickly.  
However, it takes a long time to get an initial solution, because it has to
samples IK solutions at all the end-effector poses
(Table~\ref{tab:result}).  These results show that Stampede and EIGS have
different pros and cons.  Nevertheless, our method shows anytime property as
well as highly-accurate solutions (Fig.~\ref{fig:results_chart}).

\Skip{
	Also, Stampede has relatively
	short edges \YOON{of what?} rather than EIGS, since Stampede only connects
	nodes in two adjacent end-effector poses thanks to already finely divided
	end-effector poses; in Table~\ref{tab:result_table}, EIGS shows the highest
	joint distance
	on average for all problems, due to long edges.
	Hence, Stampede takes less time to check collision and compute the pose error \YOON{compute? what does it mean here?}
	than EIGS.  
}

RelaxedIK is a real-time planner, thus it quickly finds a
solution (Table~\ref{tab:result}).  However, its accuracy is much lower
than other methods; see Fig.~\ref{fig:results_hello} and
Fig.~\ref{fig:results_chart}.
In conclusion, RelaxedIK shows real-time performance by quickly optimizing the
joint configuration for one pose, but it is difficult to obtain a
highly-accurate trajectory.

Overall, our method finds an initial solution quite quickly with a high pose error, but improves its quality as we have more planning time
(Fig.~\ref{fig:results_chart}).  Also, our method has a lower pose error on
average than other methods, as shown in Table~\ref{tab:result}.  These results support that our optimization process efficiently reduces the pose error, while satisfying several constraints.
The max. pose errors of our method are lower than others, except Stampede.
Since our algorithm has randomness when we get IK solutions at sub-sampled
poses, its max. errors can be higher than Stampede.  
Fortunately, our method shows the smallest min. error, resulting in the best accuracy on average.
This is thanks to the fast computation of the functional gradient of our
objectives, leading to find a highly-accurate solution escaping from local
minima given the planning time.


\begin{table}[t]
	\vspace{0.25cm}
	\centering
	\scriptsize
	\renewcommand \arraystretch{1.4}
	\setlength{\tabcolsep}{1.8pt}
	\caption{ 
		Ablation study of our proposed methods.
	}
	\begin{tabular}{|c|c|c|c|c|c|c|c|c|}
		\hline
		\multicolumn{2}{|c|}{~} & \multicolumn{3}{c|}{Pose error} & ~\multirow{2}{*}{TL}~ & ~\multirow{2}{*}{IST}~ & ~\multirow{2}{*}{PT}~ \\
		\cline{3-5}
		\multicolumn{2}{|c|}{~} & Average & Min. & Max. & ~ & ~ & ~ \\
		\hline
		\hline
		Square & ~Ours w/o TSGD~ & 2.46e-3 & 2.19e-3 & 2.84e-3 & \textbf{9.2} & 5.2 & ~\multirow{3}{*}{50}~ \\
		\cline{2-7}
		tracing & Ours w/o IE & 1.67e-4 & 1.29e-5 & 2.48e-3 & 9.7 & 5.2 & ~ \\
		\cline{2-7}
		w/ obs. & Ours & \textbf{~1.91e-5~} & \textbf{~1.01e-5~} & \textbf{~7.65e-5~} & 9.5 & 5.2 &  ~ \\
		\hline
		\hline
		``S" & Ours w/o TSGD & 2.37e-3 & 1.75e-3 & 3.78e-3 & 8.1 & 5.2 & ~\multirow{3}{*}{50}~ \\
		\cline{2-7}
		tracing & Ours w/o IE & 1.56e-3 & 1.12e-5 & 2.75e-2 & \textbf{7.7} & 5.2 & ~ \\
		\cline{2-7}
		\cline{2-7}
		w/ obs. & Ours & \textbf{2.40e-5} & \textbf{6.08e-6} & \textbf{8.50e-5} & 7.8 & 5.2 & ~ \\
		\hline
		\hline
		Rotation & Ours w/o TSGD & 9.87e-3 & 1.95e-3 & 1.87e-2 & \textbf{12.1} & 2.4 & ~\multirow{3}{*}{50}~ \\
		\cline{2-7}
		task& Ours w/o IE & 8.66e-3 & 7.01e-5 & 1.65e-2 & 14.3 & 1.8 & ~ \\
		\cline{2-7}
		~w/o obs.~ & Ours & \textbf{7.67e-5} & \textbf{6.72e-5} & \textbf{9.87e-5} & 13.9 & 1.8 & ~ \\
		\hline
		\hline
		Writing & Ours w/o TSGD & 9.68e-3 & 9.25e-3 & 1.04e-2 & \textbf{25.4} & 5.3 & ~\multirow{3}{*}{50}~ \\
		\cline{2-7}
		``hello" & Ours w/o IE & 5.99e-4 & 5.07e-5 & 9.83e-3 & 27.0 & 5.2 & ~ \\
		\cline{2-7}
		~w/o obs.~ & Ours & \textbf{5.11e-5} & \textbf{4.77e-5} & \textbf{6.38e-5} & 26.4 & 5.2 & ~ \\
		\hline
	\end{tabular}
	\begin{tablenotes}
		\item[1] TL: Trajectory length (rad). IST: Initial solution time (s). PT: Planning time (s).
	\end{tablenotes}
	\vspace{-0.12cm}
	\label{tab:result_ablated}
\end{table}

\noindent
\textbf{Ablated methods for our method.}
To see the benefits of our proposed methods, we conduct ablabation study of our
method by disabling the two-stage gradient descents (TSGD) and instead updating
three functional gradients at once. We also test our methods without the
iterative exploration (IE) that just tests a single trajectory updated by
TSGD.

The results of ablated methods for our method show a similar performance trend
regardless of external obstacles. Also, all of the methods find an initial
solution almost at a similar time (Fig.~\ref{fig:results_chart}).
Excluding the TSGD has a higher pose error on average than our full method, and
also has the highest min. pose error among $20$ tests, compared to other
ablated methods (Table~\ref{tab:result_ablated}).  These results demonstrate that
the different functional gradients conflict with each other.  Therefore, the
TSGD prevents the competition of different functional gradients and is useful
to get a highly-accurate solution.

Fig.~\ref{fig:results_chart} shows that excluding the IE (the sky-blue line)
does not reduce the pose error after certain seconds (e.g., $7$ seconds in
Fig.~\ref{fig:res_hello}).  This is because it was stuck in a local minimum and
thus did not find a better solution as we have more planning time.  As a
result, excluding the IE has the largest difference between the min. and max.
pose errors, as shown in Table~\ref{tab:result_ablated}. 

\Skip{
	Table~\ref{tab:result_table} also shows excluding the WEP that has better performances compared to other ablated methods. On the other hand, excluding the WEP decreases the pose error slower than our method. 
	In most cases, we can see the difference occurring before $20$ seconds (Fig.~\ref{fig:results_chart}), 
	since excluding the WEP did not sufficiently explore new trajectories at that time.   
	This result indicates that giving more weight on exploring part in IE enhances a probability to find a better solution.
}

In conclusion, these results show that our method with the TSGD and IE
synthesizes highly-accurate trajectories, while effectively getting out of
local minima.

	\section{Conclusion and Limitations}
\label{sec:5}

In this paper, we have presented the trajectory optimization of a redundant
manipulator (TORM) that holistically incorporates three important properties
into the trajectory optimization process by integrating the
Jacobian-based IK solving method and an optimization-based motion planning
approach.  Given different properties, we have suggested the two-stage gradient
descent to follow a given end-effector path and to make a feasible trajectory.
We have also performed iterative exploration 
to avoid local minima.  We have shown the benefits of our method over the
state-of-the-art techniques in environments w/ and w/o external obstacles.  Our
method has robustly minimized the pose error in a progressive manner and
achieved the highly-accurate trajectory at a reasonable time compared to other
methods.
Further, we have verified the feasibility of our synthesized trajectory using
the real, Fetch manipulator. 

While our method finds an accurate solution reasonably fast, there is no
theoretical analysis of the error reduction rate of our approach.  While the
theoretical analysis could be challenging, it would shed light on deeply
understanding the proposed approach in various aspects. Finally, we would like
to efficiently handle dynamic environments by computing signed distance
functions in real-time.

	\section*{Acknowledgment}
	We would like to thank anonymous reviewers for constructive comments. 
	This work was supported by SW Starlab program (IITP-2015-0-00199) and NRF/MSIT (No.2019R1A2C3002833).

	{
		\small
		\bibliographystyle{ieee/ieee}
		\bibliography{./root}
	}

\end{document}